\documentclass[english]{article}
\usepackage[latin9]{inputenc}
\usepackage{array}
\usepackage{float}
\usepackage{textcomp}
\usepackage{mathrsfs}
\usepackage{multirow}
\usepackage{tablefootnote}
\usepackage{amsmath}
\usepackage{amsthm}
\usepackage{amssymb}
\usepackage{stackrel}
\usepackage{graphicx}

\makeatletter

\providecommand{\tabularnewline}{\\}
\floatstyle{ruled}
\newfloat{algorithm}{tbp}{loa}
\providecommand{\algorithmname}{Algorithm}
\floatname{algorithm}{\protect\algorithmname}

\usepackage{algorithm,algpseudocode}
\usepackage{amsmath}
\algnewcommand\algorithmicforeach{\textbf{for each}}
\algdef{S}[FOR]{ForEach}[1]{\algorithmicforeach\ #1\ \algorithmicdo}

\usepackage{iclr2020_conference}
\iclrfinalcopy 
\renewcommand{\cite}{\citep}
\usepackage{hyperref} 
\makeatother

\usepackage{babel}
\begin{document}
\title{Neural Stored-program Memory}
\author{Hung Le, Truyen Tran and Svetha Venkatesh\\
Applied AI Institute, Deakin University, Geelong, Australia\\
\texttt{\{lethai,truyen.tran,svetha.venkatesh\}@deakin.edu.au}}
\maketitle
\begin{abstract}
Neural networks powered with external memory simulate computer behaviors.
These models, which use the memory to store data for a neural controller,
can learn algorithms and other complex tasks. In this paper, we introduce
a new memory to store \emph{weights} for the controller, analogous
to the stored-program memory in modern computer architectures. The
proposed model, dubbed Neural Stored-program Memory, augments current
memory-augmented neural networks, creating differentiable machines
that can switch programs through time, adapt to variable contexts
and thus resemble the Universal Turing Machine. A wide range of experiments
demonstrate that the resulting machines not only excel in classical
algorithmic problems, but also have potential for compositional, continual,
few-shot learning and question-answering tasks. 
 
\end{abstract}

\section{Introduction}

Recurrent Neural Networks (RNNs) are Turing-complete \cite{siegelmann1995computational}.
However, in practice RNNs struggle to learn simple procedures as they
lack explicit memory \cite{graves2014neural,mozer1993connectionist}.
These findings have sparked a new research direction called Memory
Augmented Neural Networks (MANNs) that emulate modern computer behavior
by detaching memorization from computation via memory and controller
network, respectively. MANNs have demonstrated significant improvements
over memory-less RNNs in various sequential learning tasks \cite{graves2016hybrid,le2018variational,NIPS2015_5846}.
Nonetheless, MANNs have barely simulated general-purpose computers.

Current MANNs miss a key concept in computer design: stored-program
memory. The concept has emerged from the idea of Universal Turing
Machine (UTM) \cite{turing1936} and further developed in Harvard
Architecture \cite{BROESCH2009135}, Von Neumann Architecture \cite{vonNeumann:1993:FDR:612487.612553}.
In UTM, both data and programs that manipulate the data are stored
in memory. A control unit then reads the programs from the memory
and executes them with the data. This mechanism allows flexibility
to perform universal computations. Unfortunately, current MANNs such
as Neural Turing Machine (NTM) \cite{graves2014neural}, Differentiable
Neural Computer (DNC) \cite{graves2016hybrid} and Least Recently
Used Access (LRUA) \cite{santoro2016meta} only support memory for
data and embed a single program into the controller network, which
goes against the stored-program memory principle. 

Our goal is to advance a step further towards UTM by coupling a MANN
with an external program memory. The program memory co-exists with
the data memory in the MANN, providing more flexibility, reuseability
and modularity in learning complicated tasks. The program memory stores
the weights of the MANN's controller network, which are retrieved
quickly via a key-value attention mechanism across timesteps yet updated
slowly via backpropagation. By introducing a meta network to moderate
the operations of the program memory, our model, henceforth referred
to as Neural Stored-program Memory (NSM), can learn to switch the
programs/weights in the controller network appropriately, adapting
to different functionalities aligning with different parts of a sequential
task, or different tasks in continual and few-shot learning.

To validate our proposal, the NTM armed with NSM, namely Neural Universal
Turing Machine (NUTM), is tested on a variety of synthetic tasks including
algorithmic tasks from \citet{graves2014neural}, composition of algorithmic
tasks and continual procedure learning. For these algorithmic problems,
we demonstrate clear improvements of NUTM over NTM. Further, we investigate
NUTM in few-shot learning by using LRUA as the MANN and achieve notably
better results. Finally, we expand NUTM application to linguistic
problems by equipping NUTM with DNC core and achieve competitive performances
against state-of-the-arts in the bAbI task \cite{weston2015towards}. 

Taken together, our study advances neural network simulation of Turing
Machines to neural architecture for Universal Turing Machines. This
develops a new class of MANNs that can store and query both the weights
and data of their own controllers, thereby following the stored-program
principle. A set of five diverse experiments demonstrate the computational
universality of the approach.

\section{Background\label{sec:Background}}

In this section, we briefly review MANN and its relations to Turing
Machines. A MANN consists of a controller network and an external
memory $\mathbf{M}\in\mathbb{R}^{N\times M}$, which is a collection
of $N$ $M$-dimensional vectors. The controller network is responsible
for accessing the memory, updating its state and optionally producing
output at each timestep. The first two functions are executed by an
interface network and a state network\footnote{Some MANNs (e.g., NTM with Feedforward Controller) neglect the state
network, only implementing the interface network and thus analogous
to one-state Turing Machine. }, respectively. Usually, the interface network is a Feedforward neural
network whose input is $c_{t}$ - the output of the state network
implemented as RNNs. Let $W^{c}$ denote the weight of the interface
network, then the state update and memory control are as follows,

\begin{minipage}[t]{0.5\textwidth}%
\begin{equation}
h_{t},c_{t}=RNN\left(\left[x_{t},r_{t-1}\right],h_{t-1}\right)
\end{equation}
\end{minipage}%
\begin{minipage}[t]{0.5\textwidth}%
\begin{equation}
\xi_{t}=c_{t}W^{c}
\end{equation}
\end{minipage}

where $x_{t}$ and $r_{t-1}$ are data from current input and the
previous memory read, respectively. The interface vector $\xi_{t}$
then is used to read from and write to the memory $\mathbf{M}$. We
use a generic notation $memory\left(\xi_{t},\mathbf{M}\right)$ to
represent these memory operations that either update or retrieve read
value $r_{t}$ from the memory. To support multiple memory accesses
per step, the interface network may produce multiple interfaces, also
known as control heads. Readers are referred to App. \ref{subsec:Example-of-memory}
and \citet{graves2014neural,graves2016hybrid,santoro2016meta} for
details of memory read/write examples. 

A deterministic one-tape Turing Machine can be defined by 4-tuple
$\left(Q,\Gamma,\delta,q_{0}\right)$, in which $Q$ is finite set
of states, $q_{0}\in Q$ is an initial state, $\Gamma$ is finite
set of symbol stored in the tape (the data) and $\delta$ is the transition
function (the program), $\delta:Q\times\Gamma\rightarrow\Gamma\times\left\{ -1,1\right\} \times Q$.
At each step, the machine performs the transition function, which
takes the current state and the read value from the tape as inputs
and outputs actions including writing new values, moving tape head
to new location (left/right) and jumping to another state. Roughly
mapping to current MANNs, $Q$, $\Gamma$ and $\delta$ map to the
set of the controller states, the read values and the controller network,
respectively. Further, the function $\delta$ can be factorized into
two sub functions: $Q\times\Gamma\rightarrow\Gamma\times\left\{ -1,1\right\} $
and $Q\times\Gamma\rightarrow Q$, which correspond to the interface
and state networks, respectively. 

By encoding a Turing Machine into the tape, one can build a UTM that
simulates the encoded machine \cite{turing1936}. The transition function
$\delta_{u}$ of the UTM queries the encoded Turing Machine that solves
the considering task. Amongst 4 tuples, $\delta$ is the most important
and hence uses most of the encoding bits. In other words, if we assume
that the space of $Q$, $\Gamma$ and $q_{0}$ are shared amongst
Turing Machines, we can simulate any Turing Machine by encoding only
its transition function $\delta$. Translating to neural language,
if we can store the controller network into a queriable memory and
make use of it, we can build a Neural Universal Turing Machine. Using
NSM is a simple way to achieve this goal, which we introduce in the
subsequent section.

\section{Methods}

\subsection{Neural Stored-program Memory}

A Neural Stored-program Memory (NSM) is a key-value memory $\mathbf{M}_{p}\in\mathbb{R}^{P\times(K+S)}$,
whose values are the basis weights of another neural network$-$the
programs. $P$, $K$, and $S$ are the number of programs, the key
space dimension and the program size, respectively. This concept is
a hybrid between the traditional slow-weight and fast-weight \cite{hinton1987using}.
Like slow-weight, the keys and values in NSM are updated gradually
by backpropagation. However, the values are dynamically interpolated
to produce the working weight on-the-fly during the processing of
a sequence, which resembles fast-weight computation. Let us denote
$\mathbf{M}_{p}\left(i\right).k$ and $\mathbf{M}_{p}\left(i\right).v$
as the key and the program of the $i$-th memory slot. At timestep
$t,$ given a query key $k_{t}^{p}$, the working program is retrieved
as follows,

\begin{equation}
D\left(k_{t}^{p},\mathbf{M}_{p}(i).k\right)=\frac{k_{t}^{p}\cdot\mathbf{M}_{p}(i).k}{||k_{t}^{p}||\cdot||\mathbf{M}_{p}(i).k)||}\label{eq:d_p}
\end{equation}

\begin{equation}
w_{t}^{p}\left(i\right)=\text{softmax}\left(\beta_{t}^{p}D\left(k_{t}^{p},\mathbf{M}_{p}(i).k\right)\right)\label{eq:pt-1}
\end{equation}

\begin{equation}
p_{t}=\stackrel[i=1]{P}{\sum}w_{t}^{p}\left(i\right)\mathbf{M}_{p}\left(i\right).v\label{eq:pt}
\end{equation}
where $D\left(\cdot\right)$ is cosine similarity and $\beta_{t}^{p}$
is the scalar program strength parameter. The vector working program
$p_{t}$ is then reshaped to its matrix form and ready to be used
as the weight of other neural networks. 

The key-value design is essential for convenient memory access as
the size of the program stored in $\mathbf{M}_{p}$ can be millions
of dimensions and thus, direct content-based addressing as in \citet{graves2014neural,graves2016hybrid,santoro2016meta}
is infeasible. More importantly, we can inject external control on
the behavior of the memory by imposing constraints on the key space.
For example, program collapse will happen when the keys stored in
the memory stay close to each other. When this happens, $p_{t}$ is
a balanced mixture of all programs regardless of the query key and
thus having multiple programs is useless. We can avoid this phenomenon
by minimizing a regularization loss defined as the following,

\begin{equation}
l_{p}=\stackrel[i=1]{P}{\sum}\stackrel[j=i+1]{P}{\sum}D\left(\mathbf{M}_{p}(i).k,\mathbf{M}_{p}(j).k\right)\label{eq:l_p}
\end{equation}

\subsection{Neural Universal Turing Machine\label{subsec:Neural-Universal-Turing}}

It turns out that the combination of MANN and NSM approximates a Universal
Turing Machine (Sec. \ref{sec:Background}). At each timestep, the
controller in MANN reads its state and memory to generate control
signal to the memory via the interface network $W^{c}$, then updates
its state using the state network $RNN$. Since the parameters of
$RNN$ and $W^{c}$ represent the encoding of $\delta$, we should
store both into NSM to completely encode an MANN. For simplicity,
in this paper, we only use NSM to store $W^{c}$, which is equivalent
to the Universal Turing Machine that can simulate any one-state Turing
Machine. 

\begin{figure}
\begin{centering}
\includegraphics[width=0.9\columnwidth]{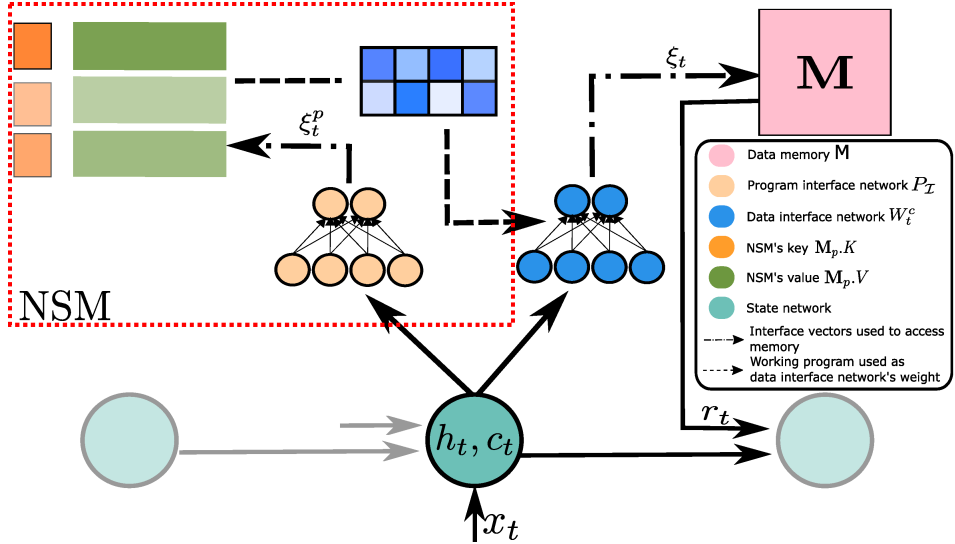}
\par\end{centering}
\caption{Introducing NSM into MANN. At each timestep, the program interface
network ($P_{\mathscr{\mathcal{I}}}$) receives input from the state
network and queries the program memory $\mathbf{M}_{p}$, acquiring
the working weight for the interface network ($W_{t}^{c}$). The interface
network then operates on the data memory $\mathbf{M}$. \label{fig:NUTM-diagram}}
\end{figure}

In traditional MANN, $W^{c}$ is constant across timesteps and only
updated slowly during training, typically through backpropagation.
In our design, we compute $W_{t}^{c}$ from NSM for every timestep
and thus, we need a program interface network$-$the meta network
$P_{\mathscr{\mathcal{I}}}-$that generates an interface vector for
the program memory: $\xi_{t}^{p}=P_{\mathscr{\mathcal{I}}}\left(c_{t}\right)$,
where $\xi_{t}^{p}=\left[k_{t}^{p},\beta_{t}^{p}\right]$. Together
with the $RNN$, $P_{\mathscr{\mathcal{I}}}$ simulates $\delta_{u}$
of the UTM and is implemented as a Feedforward neural network. The
procedure for computing $W_{t}^{c}$ is executed by following Eqs.
(\ref{eq:d_p})-(\ref{eq:pt}), hereafter referred to as $NSM\left(\xi_{t}^{p},\mathbf{M}_{p}\right)$.
Figure \ref{fig:NUTM-diagram} depicts the integration of NSM into
MANN.

In this implementation, key-value NSM offers a more flexible learning
scheme than direct attention, in which the meta-network can generate
the weight $w_{t}^{p}$ directly without matching $k_{t}^{p}$ with
$\mathbf{M}_{p}\left(i\right).k$. That is, only the meta-network
learns the mapping from context $c_{t}$ to program. When it falls
into some local-minima (generating suboptimal $w_{t}^{p}$), the meta-network
struggles to escape. In our proposal, together with the meta-network,
the memory keys are learnable. When the memory keys are slowly updated,
the meta-network will shift its query key generation to match the
new memory keys and possibly escape from the local-minima. 

For the case of multi-head NTM, we implement one NSM per control head
and name this model Neural Universal Turing Machine (NUTM). One NSM
per head is to ensure programs for one head do not interfere with
other heads and thus, encourage functionality separation amongst heads.
Each control head will read from (for read head) or write to (for
write head) the data memory $\mathbf{M}$ via $memory\left(\xi_{t},\mathbf{M}\right)$
as described in \citet{graves2014neural} . It should be noted that
using multiple heads is unlike using multiple controllers per head.
The former increases the number of accesses to the data memory at
each timestep and employs a fixed controller to compute multiple heads,
which may improve capacity yet does not enable adaptability. On the
contrary, the latter varies the property of each memory access across
timesteps by switching the controllers and thus potential for adaptation. 

Other MANNs such as DNC \cite{graves2016hybrid} and LRUA \cite{santoro2016meta}
can be armed with NSM in this manner. We also employ the regularization
loss $l_{p}$ to prevent the programs from collapsing, resulting in
a final loss as follows,

\begin{equation}
Loss=Loss_{pred}+\eta_{t}l_{p}\label{eq:loss}
\end{equation}
where $Loss_{pred}$ is the prediction loss and $\eta_{t}$ is annealing
factor, reducing as the training step increases. The details of NUTM
operations are presented in Algorithm \ref{alg:Neural-Uinversal-Turing}. 

\begin{algorithm}[t]
\begin{algorithmic}[1]
\Require{a sequence $x=\left\{ x_{t}\right\} _{t=1}^{T}$, a data memory $\mathbf{M}$ and $R$ program memories $\left\{ \mathbf{M}_{p,n}\right\} _{n=1}^{R}$ corresponding to $R$ control heads}
\State{Initilize $h_0$, $r_0$}
\For{$t=1,T$}
\State{$h_t,c_t=RNN([x_t,r_{t-1}],h_{t-1})$} \Comment{$RNN$ can be replaced by GRU/LSTM}
\For{$n=1,R$}
\State{Compute the program interface $\xi_{t,n}^{p}\leftarrow P_{\mathscr{\mathcal{I}},n}\left(c_{t}\right)$}
\State{Compute the program $W_{t,n}^{c}\leftarrow NSM\left(\xi_{t,n}^{p},\mathbf{M}_{p,n}\right)$}
\State{Compute the data interface $\xi_{t,n}\leftarrow c_{t}W_{t,n}^{c}$}
\State{\parbox[t]{0.9\linewidth}{Read $r_{t,n}$ from memory $\mathbf{M}$ (if read head) or update memory $\mathbf{M}$ (if write head) using $memory_n(\xi_{t,n},\mathbf{M})$}}
\EndFor
\State{$r_{t}\leftarrow\left[r_{t,1},...,r_{t,R}\right]$}
\EndFor
\end{algorithmic} 

\caption{Neural Universal Turing Machine\label{alg:Neural-Uinversal-Turing}}
\end{algorithm}

\subsection{On the Benefit of NSM to MANN: An Explanation from Multilevel Modeling\label{subsec:On-the-Benefit}}

Learning to access memory is a multi-dimensional regression problem.
Given the input $c_{t}$, which is derived from the state $h_{t}$
of the controller, the aim is to generate a correct interface vector
$\xi_{t}$ via optimizing the interface network. Instead of searching
for one transformation that maps the whole space of $c_{t}$ to the
optimal space of $\xi_{t}$, NSM first partitions the space of $c_{t}$
into subspaces, then finds multiple transformations, each of which
covers subspace of $c_{t}$. The program interface network $P_{\mathscr{\mathcal{I}}}$
is a meta learner that routes $c_{t}$ to the appropriate transformation,
which then maps $c_{t}$ to the $\xi_{t}$ space. This is analogous
to multilevel regression in statistics \cite{andrew2006mr}. Practical
studies have shown that multilevel regression is better than ordinary
regression if the input is clustered \cite{cohen2014applied,huang2018multilevel}. 

RNNs have the capacity to learn to perform finite state computations
\cite{casey1996dynamics,tivno1998finite}. The states of a RNN must
be grouped into partitions representing the states of the generating
automaton. As Turing Machines are finite state automata augmented
with an external memory tape, we expect MANN, if learnt well, will
organize its state space clustered in a way to reflect the states
of the emulated Turing Machine. That is, $h_{t}$ as well as $c_{t}$
should be clustered. We realize that NSM helps NTM learn better clusterization
over this space (see App. \ref{subsec:Clustering-on-The}), thereby
improving NTM's performances.

\section{Results}

\subsection{NTM Single Tasks\label{subsec:NTM-Single-Tasks}}

\begin{figure}
\begin{centering}
\includegraphics[width=1\textwidth]{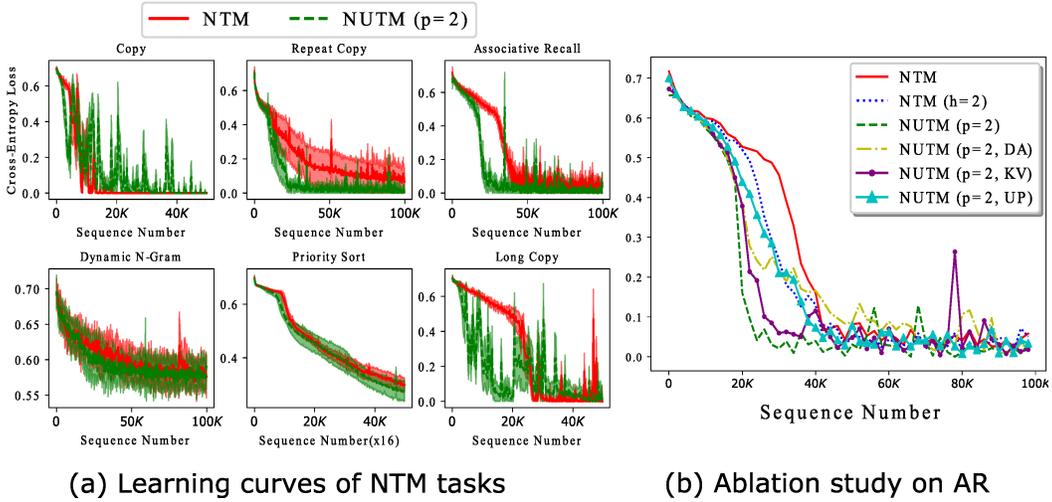}
\par\end{centering}
\caption{Learning curves on NTM tasks (a) and Associative Recall (AR) ablation
study (b). Only mean is plotted in (b) for better visualization.\label{fig:Learning-curves-on}}
\end{figure}
\begin{table}
\begin{centering}
\begin{tabular}{c|cccccc}
\hline 
Task & Copy & R. Copy & A. Recall & D. N-grams & P. Sort & L. Copy\tabularnewline
\hline 
NTM & \textbf{0.00} & 405.10 & 7.66 & 132.59 & 24.41 & 16.04\tabularnewline
NUTM (p=2) & \textbf{0.00} & \textbf{366.69} & \textbf{1.35} & \textbf{127.68} & \textbf{20.00} & \textbf{0.02}\tabularnewline
\hline 
\end{tabular}
\par\end{centering}
~

\caption{Generalization performance of best models measured in average bit
error per sequence (lower is better). For each task, we pick 1,000
longer sequences as test data. \label{tab:Generalisation-performance-of}}
\end{table}
In this section, we investigate the performance of NUTM on algorithmic
tasks introduced in \citet{graves2014neural} : Copy, Repeat Copy,
Associative Recall, Dynamic N-Grams and Priority Sort. Besides these
five NTM tasks, we add another task named Long Copy which doubles
the length of training sequences in the Copy task. In these tasks,
the model will be fed a sequence of input items and is required to
infer a sequence of output items. Each item is represented by a binary
vector.

In the experiment, we compare two models: NTM\footnote{For algorithmic tasks, we choose NTM as the only baseline as NTM is
known to perform and generalize well on these tasks. If NSM can help
NTM in these tasks, it will probably help other MANNs as well.} and NUTM with two programs. Although the tasks are atomic, we argue
that there should be at least two memory manipulation schemes across
timesteps, one for encoding the inputs to the memory and another for
decoding the output from the memory. The two models are trained with
cross-entropy objective function under the same setting as in \citet{graves2014neural}
. For fair comparison, the controller hidden dimension of NUTM is
set smaller to make the total number of parameters of NUTM equivalent
to that of NTM. The number of memory heads for both models are always
equal and set to the same value as in the original paper (details
in App. \ref{subsec:Details-on-Synthetic}).

We run each experiments five times and report the mean with error
bars of training losses for NTM tasks in Fig. \ref{fig:Learning-curves-on}
(a). Except for the Copy task, which is too simple, other tasks observe
convergence speed improvement of NUTM over that of NTM, thereby validating
the benefit of using two programs across timesteps even for the single
task setting. NUTM requires fewer training samples to converge and
it generalizes better to unseen sequences that are longer than training
sequences. Table \ref{tab:Generalisation-performance-of} reports
the test results of the best models chosen after five runs and confirms
the outperformance of NUTM over NTM for generalization. 

To illustrate the program usage, we plot NUTM's program distributions
across timesteps for Repeat Copy and Priority Sort in Fig. \ref{fig:Memory-read-(a,c,d)/write}
(a) and (b), respectively. Examining the read head for Repeat Copy,
we observe two program usage patterns corresponding to the encoding
and decoding phases. As there is no reading in encoding, NUTM assigns
the ``no-read'' strategy mainly to the ``orange program''. In
decoding, the sequential reading is mostly done by the ``blue program''
with some contributions from the ``orange program'' when resetting
reading head. Similar behaviors can be found in the write head for
Priority Sort. While the encoding ``fitting writing'' (see \citet{graves2014neural}
for explanation on the strategy) is often executed by the ``blue
program'', the decoding writing is completely taken by the ``orange''
program (more visualizations in App. \ref{subsec:Program-Usage-Visualizations}). 

\begin{figure}
\begin{centering}
\includegraphics[width=0.95\linewidth]{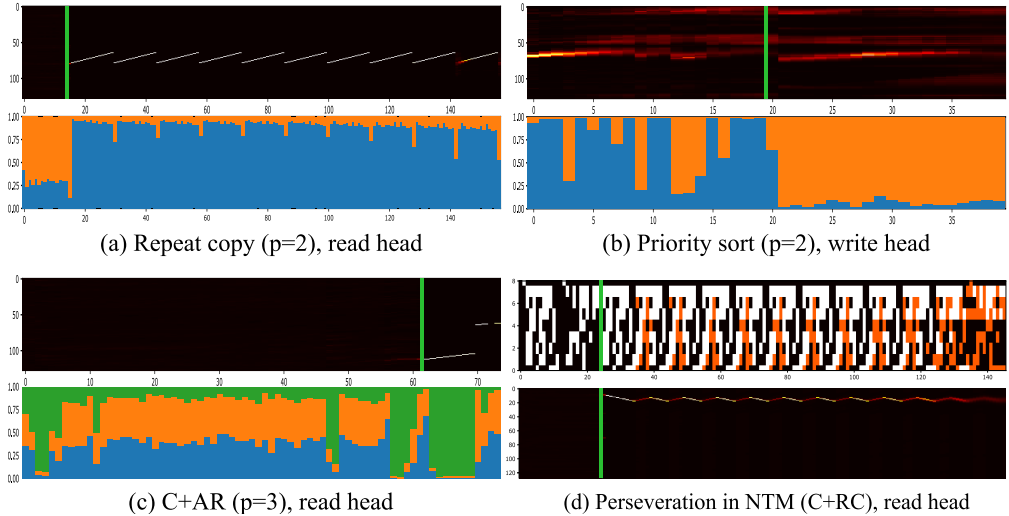}
\par\end{centering}
\caption{(a,b,c) visualizes NUTM's executions in synthetic tasks: the upper
rows are memory read (left)/write (right) locations; the lower rows
are program distributions over timesteps. The green line indicates
the start of the decoding phase. (d) visualizes perseveration in NTM:
the upper row are input, output, predicted output with errors (orange
bits); the lower row is reading location. \label{fig:Memory-read-(a,c,d)/write}}
\end{figure}

\subsection{Ablation study on Associative Recall}

In this section, we conduct an ablation study on Associative Recall
(AR) to validate the benefit of proposed components that constitute
NSM. We run the task with three additional baselines: NUTM using direct
attention (DA), NUTM using key-value without regularization (KV),
NUTM using fixed, uniform program distribution (UP) and a vanilla
NTM with 2 memory heads ($h=2$). The meta-network $P_{\mathscr{\mathcal{I}}}$
in DA generates the attention weight $w_{t}^{p}$ directly. The KV
employs key-value attention yet excludes the regularization loss presented
in Eq. (\ref{eq:l_p}). The training curves over 5 runs are plotted
in Fig. \ref{fig:Learning-curves-on} (b). The results demonstrate
that DA exhibits fast yet shallow convergence. It tends to fall into
local minima, which finally fails to reach zero loss. Key-value attention
helps NUTM converge completely with fewer iterations. The performance
is further improved with the proposed regularization loss. UP underperforms
NUTM as it lacks dynamic programs. The NTM with 2 heads shows slightly
better convergence compared to the NTM, yet obviously underperforms
NUTM ($p=2$) with 1 head and fewer parameters. This validates our
argument on the difference between using multiple heads and multiple
programs (Sec. \ref{subsec:Neural-Universal-Turing}).

\subsection{NTM Sequencing Tasks}

In neuroscience, sequencing tasks test the ability to remember a series
of tasks and switch tasks alternatively \cite{hal2019neur}. A dysfunctional
brain may have difficulty in changing from one task to the next and
get stuck in its preferred task (perseveration phenomenon). To analyze
this problem in NTM, we propose a new set of experiments in which
a task is generated by sequencing a list of subtasks. The set of subtasks
is chosen from the NTM single tasks (excluding Dynamic N-grams for
format discrepancy) and the order of subtasks in the sequence is dictated
by an indicator vector put at the beginning of the sequence. Amongst
possible combinations of subtasks, we choose \{Copy, Repeat Copy\}(C+RC),
\{Copy, Associative Recall\} (C+AR), \{Copy, Priority Sort\} (C+PS)
and all (C+RC+AC+PS)\footnote{We focus on the combinations that contain Copy as Copy is the only
task where NTM reach NUTM's performance. If NTM fails in these combinations,
it will most likely fail in others.}. The learner observes the order indicator followed by a sequence
of subtasks' input items and is requested to consecutively produce
the output items of each subtasks. 

\begin{figure}
\begin{centering}
\includegraphics[width=0.95\linewidth]{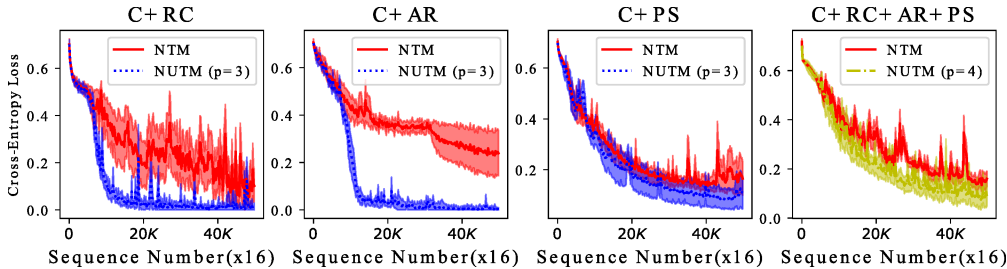}
\par\end{centering}
\caption{Learning curves on sequencing NTM tasks.\label{fig:Learning-curves-on-1}}
\end{figure}

As shown in Fig. \ref{fig:Learning-curves-on-1}, some tasks such
as Copy and Associative Recall, which are easy to solve if trained
separately, become unsolvable by NTM when sequenced together. One
reason is NTM fails to change the memory access behavior (perseveration).
For examples, NTM keeps following repeat copy reading strategy for
all timesteps in C+RC task (Fig. \ref{fig:Memory-read-(a,c,d)/write}
(d)). Meanwhile, NUTM can learn to change program distribution when
a new subtask appears in the sequence and thus ensure different accessing
strategy per subtask (Fig. \ref{fig:Memory-read-(a,c,d)/write} (c)).

\subsection{Continual Procedure Learning}

In continual learning, catastrophic forgetting happens when a neural
network quickly forgets previously acquired skills upon learning new
skills \cite{french1999catastrophic}. In this section, we prove the
versatility of NSM by showing that a naive application of NSM without
much modification can help NTM to mitigate catastrophic forgetting.
We design an experiment similar to the Split MNIST \cite{zenke2017continual}
to investigate whether NSM can improve NTM's performance. In our experiment,
we let the models see the training data from the 4 tasks: Copy (C),
Repeat Copy (RC), Associative Recall (AR) and Priority Sort (PS),
consecutively in this order. Each task is trained in 20,000 iterations
with batch size 16 (see App. \ref{subsec:Details-on-Synthetic} for
task details). To encourage NUTM to spend exactly one program per
task while freezing others, we force ``hard'' attention over the
programs by replacing the softmax function in Eq. \ref{eq:pt} with
the Gumbel-softmax \cite{jang2016categorical}. Also, to ignore catastrophic
forgetting in the state network, we use Feedforward controllers in
the two baselines.

\begin{figure}
\centering{}\includegraphics[width=0.95\linewidth]{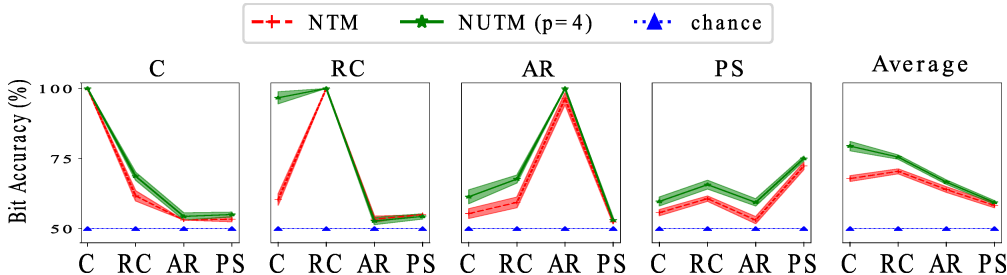}\caption{Mean bit accuracy with error bars for the continual algorithmic tasks.
Each of the first four panels show bit accuracy on four tasks after
finishing a task. The rightmost shows the average accuracy.\label{fig:Mean-bit-accuracy}}
\end{figure}

After finishing one task, we evaluate the bit accuracy $-$measured
by $1-$(bit error per sequence/total bits per sequence) over 4 tasks.
As shown in in Fig. \ref{fig:Mean-bit-accuracy}, NUTM outperforms
NTM by a moderate margin (10-40\% per task). Although NUTM also experiences
catastrophic forgetting, it somehow preserves some memories of previous
tasks. Especially, NUTM keeps performing perfectly on Copy even after
it learns Repeat Copy. For other dissimilar task transitions, the
performance drops significantly, which requires more effort to bring
NSM to continual learning. 

\subsection{Few-shot Learning}

Few-shot learning or meta learning tests the ability to rapidly adapt
within a task while gradually capturing the way the task structure
varies \cite{thrun1998lifelong}. By storing sample-class bindings,
MANNs are capable of classifying new data after seeing only few samples
 \cite{santoro2016meta}. As NSM gives flexible memory controls, it
makes MANN more adaptive to changes and thus perform better in this
setting. To verify that, we apply NSM to the LRUA memory and follow
the experiments introduced in \citet{santoro2016meta} , using the
Omniglot dataset to measure few-shot classification accuracy. The
dataset includes images of 1623 characters, with 20 examples of each
character. During training, a sequence (episode) of images are randomly
selected from $C$ classes of characters in the training set (1200
characters), where $C=5,10$ corresponding to sequence length of 50,
75, respectively. Each class is assigned a random label which shuffles
between episodes and is revealed to the models after each prediction.
After 100,000 episodes of training, the models are tested with unseen
images from the testing set (423 characters). The two baselines are
MANN and NUTM (both use LRUA core). For NUTM, we only tune $p$ and
pick the best values: $p=2$ and $p=3$ for 5 classes and 10 classes,
respectively. 

Table \ref{tab:meta} reports the classification accuracy when the
models see characters for the second, third and fifth time. NUTM generally
achieves better results than MANN, especially when the number of classes
increases, demanding more adaptation within an episode. For the persistent
memory mode, which demands fast forgetting old experiences in previous
episodes, NUTM outperforms MANN significantly (10-20\%)\footnote{It should be noted that our goal was not to achieve state of the art
performance on this dataset. It was to exhibit the benefit of NSM
to MANN. Compared to current methods, the MANN and NUTM used in our
experiments do not use CNN to extract visual features, thus achieve
lower accuracy than recent state-of-the-arts. }. Readers are referred to App. \ref{subsec:Details-on-Few-shot} for
more details on learning curves and more results of the models. 

\begin{table}
\begin{centering}
\begin{tabular}{l|c|ccc|ccc}
\hline 
\multirow{2}{*}{Model} & Persistent & \multicolumn{3}{c|}{5 classes} & \multicolumn{3}{c}{10 classes}\tabularnewline
 & memory\tablefootnote{If the memory is not artificially erased between episodes, it is called
persistent. This mode is hard for the case of 5 classes as shown in
\cite{santoro2016meta} } & $2^{nd}$ & $3^{rd}$ & $5^{th}$ & $2^{nd}$ & $3^{rd}$ & $5^{th}$\tabularnewline
\hline 
MANN (LRUA){*} & No & 82.8 & 91.0 & 94.9 & - & - & -\tabularnewline
MANN (LRUA) & No & 82.3 & 88.7 & 92.3 & 52.7 & 60.6 & 64.7\tabularnewline
NUTM (LRUA) & No & \textbf{85.7} & \textbf{91.3} & \textbf{95.5} & \textbf{68.0} & \textbf{78.1} & \textbf{82.8}\tabularnewline
\hline 
MANN (LRUA) & Yes & 66.2 & 73.4 & 81.0 & 51.3 & 59.2 & 63.3\tabularnewline
NUTM (LRUA) & Yes & \textbf{77.8} & \textbf{85.8} & \textbf{89.8} & \textbf{69.0} & \textbf{77.9} & \textbf{82.7}\tabularnewline
\hline 
\end{tabular}
\par\end{centering}
~

\caption{Test-set classification accuracy (\%) on the Omniglot dataset after
100,000 episodes of training. {*} denotes available results from \cite{santoro2016meta}.}
\end{table}

\subsection{Text Question Answering }

Reading comprehension typically involves an iterative process of multiple
actions such as reading the story, reading the question, outputting
the answers and other implicit reasoning steps \cite{weston2015towards}.
We apply NUTM to the question answering domain by replacing the NTM
core with DNC \cite{graves2016hybrid}. Compared to NTM's sequential
addressing, dynamic memory addressing in DNC is more powerful and
thus suitable for NSM integration to solve non-algorithmic problems
such as question answering. Following previous works of DNC, we use
bAbI dataset \cite{weston2015towards} to measure the performance
of the NUTM with DNC core (three variants $p=1$, $p=2$ and $p=4$).
In the dataset, each story is followed by a series of questions and
the network reads all word by word, then predicts the answers. Although
synthetically generated, bAbI is a good benchmark that tests 20 aspects
of natural language reasoning including complex skills such as induction
and counting, 

We found that increasing number of programs helps NUTM improve performance.
In particular, NUTM with 4 programs, after 50 epochs jointly trained
on all 20 question types, can achieve a mean test error rate of 3.3\%
and manages to solve 19/20 tasks (a task is considered solved if its
error \textless 5\%). The mean and s.d. across 10 runs are also compared
with other results reported by recent works (see Table \ref{tab:Mean-bAbI-error}).
Excluding baselines under different setups, our result is the best
reported mean result on bAbI that we are aware of. More details are
described in App. \ref{subsec:Details-on-bAbI}. 

\begin{table}
\begin{centering}
\begin{tabular}{ll}
\hline 
\multicolumn{1}{l}{Model} & Error\tabularnewline
\hline 
\multicolumn{1}{l}{DNC\cite{graves2016hybrid}} & 16.7 \textpm{} 7.6 \tabularnewline
\multicolumn{1}{l}{SDNC\cite{rae2016scaling}} & 6.4 \textpm{} 2.5 \tabularnewline
\multicolumn{1}{l}{ADNC\cite{W18-2606}} & 6.3 \textpm{} 2.7 \tabularnewline
\multicolumn{1}{l}{DNC-MD\cite{csordas2018improving}} & 9.5 \textpm{} 1.6\tabularnewline
\hline 
NUTM (DNC core, p=1) & 9.7 \textpm{} 3.5\tabularnewline
NUTM (DNC core, p=2) & 7.5 \textpm{} 1.6\tabularnewline
NUTM (DNC core, p=4) & \textbf{5.6 \textpm{} 1.9}\tabularnewline
\hline 
\end{tabular}
\par\end{centering}
~

\caption{Mean and s.d. for bAbI error ($\%$).\label{tab:Mean-bAbI-error}}

\end{table}

\section{Related Work}

Previous investigations into MANNs mostly revolve around memory access
mechanisms. The works in \citet{graves2014neural,graves2016hybrid}
introduce content-based, location-based and dynamic memory reading/writing.
Further, \citet{rae2016scaling} scales to bigger memory by sparse
access; \citet{le2018learning} optimizes memory operations with uniform
writing; and MANNs with extra memory have been proposed \cite{Le:2018:DMN:3219819.3219981}.
However, these works keep using memory for storing data rather than
the weights of the network and thus parallel to our approach. Other
DNC modifications \cite{csordas2018improving,W18-2606} are also orthogonal
to our work. 

Another line of related work involves modularization of neural networks,
which is designed for visual question answering. In module networks
\cite{andreas2016neural,andreas-etal-2016-learning}, the modules
are manually aligned with predefined concepts and the order of execution
is decided by the question. Although the module in these works resembles
the program in NSM, our model is more generic and flexible with soft-attention
over programs and thus fully differentiable. Further, the motivation
of NSM does not limit to a specific application. Rather, NSM aims
to help MANN reach general-purpose computability. 

If we view NSM network as a dynamic weight generator, the program
in NSM can be linked to fast weight \cite{cogprints1380,hinton1987using,schmidhuber1993self}.
These papers share the idea of using different weights across timesteps
to enable dynamic adaptation. Using outer-product is a common way
to implement fast-weight \cite{schmidhuber1993reducing,ba2016using,schlag2017gated}.
These fast weights are directly generated and thus different from
our programs, which are interpolated from a set of slow weights. 

Tensor/Multiplicative RNN \cite{sutskever2011generating} and Hypernetwork
\cite{ha2016hypernetworks} are also relevant related works. These
methods attempt to make the working weight of RNNs dependent on the
input to enable quick adaption through time. Nevertheless, they do
not support modularity. In particular, Hypernetwork generates scaling
factors for the single weight of the main RNN. It does not aim to
use multiple slow-weights (programs) and thus, different from our
approach. Tensor RNN is closer to our idea when the authors propose
to store $M$ slow-weights, where $M$ is the number of input dimension,
which is acknowledged impractical. Unlike our approach, they do not
use a meta-network to generate convex combinations amongst weights.
Instead, they propose Multiplicative RNN that factorizes the working
weight to product of three matrices, which looses modularity. On the
contrary, we explicitly model the working weight as an interpolation
of multiple programs and use a meta-network to generate the coefficients.
This design facilitates modularity because each program is trained
towards some functionality and can be switched or combined with each
other to perform the current task. Last but not least, while the related
works focus on improving RNN with fast-weight, we aim to reach a neural
simulation of Universal Turing Machine, in which fast-weight is a
way to implement stored-program principle.

\section{Conclusions}

This paper introduces the Neural Stored-program Memory (NSM), a new
type of external memory for neural networks. The memory, which takes
inspirations from the stored-program memory in computer architecture,
gives memory-augmented neural networks (MANNs) flexibility to change
their control programs through time while maintaining differentiability.
The mechanism simulates modern computer behavior, potential making
MANNs truly neural computers. Our experiments demonstrated that when
coupled with our model, the Neural Turing Machine learns algorithms
better and adapts faster to new tasks at both sequence and sample
levels. When used in few-shot learning, our method helps MANN as well.
We also applied the NSM to the Differentiable Neural Computer and
observed a significant improvement, reaching the state-of-the-arts
in the bAbI task. Although this paper limits to MANN integration,
other neural networks can also reap benefits from our proposed model,
which will be explored in future works.

\bibliographystyle{iclr2020_conference}
\bibliography{nva}

\begin{thebibliography}{37}
\providecommand{\natexlab}[1]{#1}
\providecommand{\url}[1]{\texttt{#1}}
\expandafter\ifx\csname urlstyle\endcsname\relax
  \providecommand{\doi}[1]{doi: #1}\else
  \providecommand{\doi}{doi: \begingroup \urlstyle{rm}\Url}\fi

\bibitem[Andreas et~al.(2016{\natexlab{a}})Andreas, Rohrbach, Darrell, and
  Klein]{andreas-etal-2016-learning}
Jacob Andreas, Marcus Rohrbach, Trevor Darrell, and Dan Klein.
\newblock Learning to compose neural networks for question answering.
\newblock In \emph{Proceedings of the 2016 Conference of the North {A}merican
  Chapter of the Association for Computational Linguistics: Human Language
  Technologies}, pp.\  1545--1554, 2016{\natexlab{a}}.
\newblock \doi{10.18653/v1/N16-1181}.
\newblock URL \url{https://www.aclweb.org/anthology/N16-1181}.

\bibitem[Andreas et~al.(2016{\natexlab{b}})Andreas, Rohrbach, Darrell, and
  Klein]{andreas2016neural}
Jacob Andreas, Marcus Rohrbach, Trevor Darrell, and Dan Klein.
\newblock Neural module networks.
\newblock In \emph{Proceedings of the IEEE Conference on Computer Vision and
  Pattern Recognition}, pp.\  39--48, 2016{\natexlab{b}}.

\bibitem[Andrew~Gelman(2006)]{andrew2006mr}
Jennifer~Hill Andrew~Gelman.
\newblock \emph{Data Analysis Using Regression and Multilevel/Hierarchical
  Models}.
\newblock Cambridge University Press, 2006.

\bibitem[Ba et~al.(2016)Ba, Hinton, Mnih, Leibo, and Ionescu]{ba2016using}
Jimmy Ba, Geoffrey~E Hinton, Volodymyr Mnih, Joel~Z Leibo, and Catalin Ionescu.
\newblock Using fast weights to attend to the recent past.
\newblock In \emph{Advances in Neural Information Processing Systems}, pp.\
  4331--4339, 2016.

\bibitem[Blumenfeld(2010)]{hal2019neur}
Hal Blumenfeld.
\newblock \emph{Neuroanatomy through Clinical Cases}.
\newblock Oxford University Press, 2010.

\bibitem[Broesch(2009)]{BROESCH2009135}
James~D. Broesch.
\newblock Chapter 8 - digital signal processors.
\newblock In James~D. Broesch (ed.), \emph{Digital Signal Processing}, Instant
  Access, pp.\  135 -- 146. Newnes, Burlington, 2009.
\newblock ISBN 978-0-7506-8976-2.
\newblock \doi{https://doi.org/10.1016/B978-0-7506-8976-2.00008-0}.
\newblock URL
  \url{http://www.sciencedirect.com/science/article/pii/B9780750689762000080}.

\bibitem[Casey(1996)]{casey1996dynamics}
Mike Casey.
\newblock The dynamics of discrete-time computation, with application to
  recurrent neural networks and finite state machine extraction.
\newblock \emph{Neural computation}, 8\penalty0 (6):\penalty0 1135--1178, 1996.

\bibitem[Cohen et~al.(2014)Cohen, West, and Aiken]{cohen2014applied}
Patricia Cohen, Stephen~G West, and Leona~S Aiken.
\newblock \emph{Applied multiple regression/correlation analysis for the
  behavioral sciences}.
\newblock Psychology Press, 2014.

\bibitem[Csordas \& Schmidhuber(2019)Csordas and
  Schmidhuber]{csordas2018improving}
Robert Csordas and Juergen Schmidhuber.
\newblock Improving differentiable neural computers through memory masking,
  de-allocation, and link distribution sharpness control.
\newblock In \emph{International Conference on Learning Representations}, 2019.
\newblock URL \url{https://openreview.net/forum?id=HyGEM3C9KQ}.

\bibitem[Franke et~al.(2018)Franke, Niehues, and Waibel]{W18-2606}
J{\"o}rg Franke, Jan Niehues, and Alex Waibel.
\newblock Robust and scalable differentiable neural computer for question
  answering.
\newblock In \emph{Proceedings of the Workshop on Machine Reading for Question
  Answering}, pp.\  47--59. Association for Computational Linguistics, 2018.
\newblock URL \url{http://aclweb.org/anthology/W18-2606}.

\bibitem[French(1999)]{french1999catastrophic}
Robert~M French.
\newblock Catastrophic forgetting in connectionist networks.
\newblock \emph{Trends in cognitive sciences}, 3\penalty0 (4):\penalty0
  128--135, 1999.

\bibitem[Graves et~al.(2014)Graves, Wayne, and Danihelka]{graves2014neural}
Alex Graves, Greg Wayne, and Ivo Danihelka.
\newblock Neural turing machines.
\newblock \emph{arXiv preprint arXiv:1410.5401}, 2014.

\bibitem[Graves et~al.(2016)Graves, Wayne, Reynolds, Harley, Danihelka,
  Grabska-Barwi{\'n}ska, Colmenarejo, Grefenstette, Ramalho, Agapiou,
  et~al.]{graves2016hybrid}
Alex Graves, Greg Wayne, Malcolm Reynolds, Tim Harley, Ivo Danihelka, Agnieszka
  Grabska-Barwi{\'n}ska, Sergio~G{\'o}mez Colmenarejo, Edward Grefenstette,
  Tiago Ramalho, John Agapiou, et~al.
\newblock Hybrid computing using a neural network with dynamic external memory.
\newblock \emph{Nature}, 538\penalty0 (7626):\penalty0 471--476, 2016.

\bibitem[Ha et~al.(2016)Ha, Dai, and Le]{ha2016hypernetworks}
David Ha, Andrew Dai, and Quoc~V Le.
\newblock Hypernetworks.
\newblock \emph{arXiv preprint arXiv:1609.09106}, 2016.

\bibitem[Hinton \& Plaut(1987)Hinton and Plaut]{hinton1987using}
Geoffrey~E Hinton and David~C Plaut.
\newblock Using fast weights to deblur old memories.
\newblock In \emph{Proceedings of the ninth annual conference of the Cognitive
  Science Society}, pp.\  177--186, 1987.

\bibitem[Huang(2018)]{huang2018multilevel}
Francis~L Huang.
\newblock Multilevel modeling and ordinary least squares regression: how
  comparable are they?
\newblock \emph{The Journal of Experimental Education}, 86\penalty0
  (2):\penalty0 265--281, 2018.

\bibitem[Jang et~al.(2016)Jang, Gu, and Poole]{jang2016categorical}
Eric Jang, Shixiang Gu, and Ben Poole.
\newblock Categorical reparameterization with gumbel-softmax.
\newblock \emph{arXiv preprint arXiv:1611.01144}, 2016.

\bibitem[Le et~al.(2018{\natexlab{a}})Le, Tran, Nguyen, and
  Venkatesh]{le2018variational}
Hung Le, Truyen Tran, Thin Nguyen, and Svetha Venkatesh.
\newblock Variational memory encoder-decoder.
\newblock In \emph{Advances in Neural Information Processing Systems}, pp.\
  1515--1525, 2018{\natexlab{a}}.

\bibitem[Le et~al.(2018{\natexlab{b}})Le, Tran, and
  Venkatesh]{Le:2018:DMN:3219819.3219981}
Hung Le, Truyen Tran, and Svetha Venkatesh.
\newblock Dual memory neural computer for asynchronous two-view sequential
  learning.
\newblock In \emph{Proceedings of the 24th ACM SIGKDD International Conference
  on Knowledge Discovery; Data Mining}, KDD '18, pp.\  1637--1645, New York,
  NY, USA, 2018{\natexlab{b}}. ACM.
\newblock ISBN 978-1-4503-5552-0.
\newblock \doi{10.1145/3219819.3219981}.
\newblock URL \url{http://doi.acm.org/10.1145/3219819.3219981}.

\bibitem[Le et~al.(2019)Le, Tran, and Venkatesh]{le2018learning}
Hung Le, Truyen Tran, and Svetha Venkatesh.
\newblock Learning to remember more with less memorization.
\newblock In \emph{International Conference on Learning Representations}, 2019.
\newblock URL \url{https://openreview.net/forum?id=r1xlvi0qYm}.

\bibitem[Lei~Ba et~al.(2016)Lei~Ba, Kiros, and Hinton]{lei2016layer}
Jimmy Lei~Ba, Jamie~Ryan Kiros, and Geoffrey~E Hinton.
\newblock Layer normalization.
\newblock \emph{arXiv preprint arXiv:1607.06450}, 2016.

\bibitem[Mozer \& Das(1993)Mozer and Das]{mozer1993connectionist}
Michael~C Mozer and Sreerupa Das.
\newblock A connectionist symbol manipulator that discovers the structure of
  context-free languages.
\newblock In \emph{Advances in neural information processing systems}, pp.\
  863--870, 1993.

\bibitem[Rae et~al.(2016)Rae, Hunt, Danihelka, Harley, Senior, Wayne, Graves,
  and Lillicrap]{rae2016scaling}
Jack Rae, Jonathan~J Hunt, Ivo Danihelka, Timothy Harley, Andrew~W Senior,
  Gregory Wayne, Alex Graves, and Tim Lillicrap.
\newblock Scaling memory-augmented neural networks with sparse reads and
  writes.
\newblock In \emph{Advances in Neural Information Processing Systems}, pp.\
  3621--3629, 2016.

\bibitem[Santoro et~al.(2016)Santoro, Bartunov, Botvinick, Wierstra, and
  Lillicrap]{santoro2016meta}
Adam Santoro, Sergey Bartunov, Matthew Botvinick, Daan Wierstra, and Timothy
  Lillicrap.
\newblock Meta-learning with memory-augmented neural networks.
\newblock In \emph{International conference on machine learning}, pp.\
  1842--1850, 2016.

\bibitem[Schlag \& Schmidhuber(2017)Schlag and Schmidhuber]{schlag2017gated}
Imanol Schlag and J{\"u}rgen Schmidhuber.
\newblock Gated fast weights for on-the-fly neural program generation.
\newblock In \emph{NIPS Metalearning Workshop}, 2017.

\bibitem[Schmidhuber(1993{\natexlab{a}})]{schmidhuber1993reducing}
J~Schmidhuber.
\newblock Reducing the ratio between learning complexity and number of time
  varying variables in fully recurrent nets.
\newblock In \emph{International Conference on Artificial Neural Networks},
  pp.\  460--463. Springer, 1993{\natexlab{a}}.

\bibitem[Schmidhuber(1993{\natexlab{b}})]{schmidhuber1993self}
J{\"u}rgen Schmidhuber.
\newblock A self-referential weight matrix.
\newblock In \emph{International Conference on Artificial Neural Networks},
  pp.\  446--450. Springer, 1993{\natexlab{b}}.

\bibitem[Siegelmann \& Sontag(1995)Siegelmann and
  Sontag]{siegelmann1995computational}
Hava~T Siegelmann and Eduardo~D Sontag.
\newblock On the computational power of neural nets.
\newblock \emph{Journal of computer and system sciences}, 50\penalty0
  (1):\penalty0 132--150, 1995.

\bibitem[Sukhbaatar et~al.(2015)Sukhbaatar, szlam, Weston, and
  Fergus]{NIPS2015_5846}
Sainbayar Sukhbaatar, arthur szlam, Jason Weston, and Rob Fergus.
\newblock End-to-end memory networks.
\newblock In C.~Cortes, N.~D. Lawrence, D.~D. Lee, M.~Sugiyama, and R.~Garnett
  (eds.), \emph{Advances in Neural Information Processing Systems}, pp.\
  2440--2448. 2015.
\newblock URL
  \url{http://papers.nips.cc/paper/5846-end-to-end-memory-networks.pdf}.

\bibitem[Sutskever et~al.(2011)Sutskever, Martens, and
  Hinton]{sutskever2011generating}
Ilya Sutskever, James Martens, and Geoffrey~E Hinton.
\newblock Generating text with recurrent neural networks.
\newblock In \emph{Proceedings of the 28th International Conference on Machine
  Learning (ICML-11)}, pp.\  1017--1024, 2011.

\bibitem[Thrun(1998)]{thrun1998lifelong}
Sebastian Thrun.
\newblock Lifelong learning algorithms.
\newblock In \emph{Learning to learn}, pp.\  181--209. Springer, 1998.

\bibitem[Ti{\v{n}}o et~al.(1998)Ti{\v{n}}o, Horne, Giles, and
  Collingwood]{tivno1998finite}
Peter Ti{\v{n}}o, Bill~G Horne, C~Lee Giles, and Pete~C Collingwood.
\newblock Finite state machines and recurrent neural networks--automata and
  dynamical systems approaches.
\newblock In \emph{Neural networks and pattern recognition}, pp.\  171--219.
  Elsevier, 1998.

\bibitem[Turing(1936)]{turing1936}
A.M Turing.
\newblock On computable numbers, with an application to the
  entscheidungsproblem.
\newblock In \emph{Proceedings of the London Mathematical Society}, 1936.

\bibitem[von~der Malsburg(1981)]{cogprints1380}
Christoph von~der Malsburg.
\newblock The correlation theory of brain function, 1981.
\newblock URL \url{http://cogprints.org/1380/}.

\bibitem[von Neumann(1993)]{vonNeumann:1993:FDR:612487.612553}
John von Neumann.
\newblock First draft of a report on the edvac.
\newblock \emph{IEEE Ann. Hist. Comput.}, 15\penalty0 (4):\penalty0 27--75,
  October 1993.
\newblock ISSN 1058-6180.
\newblock \doi{10.1109/85.238389}.
\newblock URL \url{https://doi.org/10.1109/85.238389}.

\bibitem[Weston et~al.(2015)Weston, Bordes, Chopra, Rush, van Merri{\"e}nboer,
  Joulin, and Mikolov]{weston2015towards}
Jason Weston, Antoine Bordes, Sumit Chopra, Alexander~M Rush, Bart van
  Merri{\"e}nboer, Armand Joulin, and Tomas Mikolov.
\newblock Towards ai-complete question answering: A set of prerequisite toy
  tasks.
\newblock \emph{arXiv preprint arXiv:1502.05698}, 2015.

\bibitem[Zenke et~al.(2017)Zenke, Poole, and Ganguli]{zenke2017continual}
Friedemann Zenke, Ben Poole, and Surya Ganguli.
\newblock Continual learning through synaptic intelligence.
\newblock In \emph{Proceedings of the 34th International Conference on Machine
  Learning-Volume 70}, pp.\  3987--3995. JMLR. org, 2017.

\end{thebibliography}

\newpage{}

\section*{Appendix}

\renewcommand\thesubsection{\Alph{subsection}}

\subsection{Clustering on The Latent Space\label{subsec:Clustering-on-The}}

\begin{figure}[H]
\begin{centering}
\includegraphics[width=0.95\linewidth]{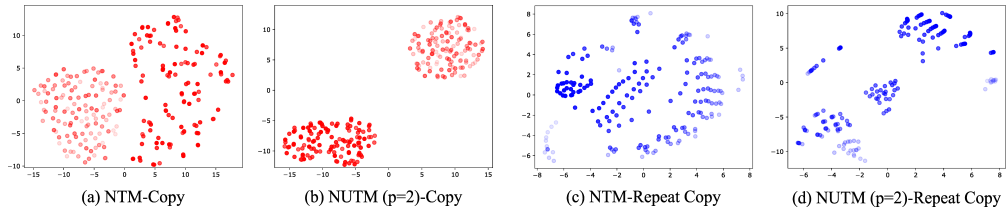}
\par\end{centering}
\caption{Visualization of the first two principal components of $c_{t}$ space
in NTM (a,c) and NUTM (b,d) for Copy (red) and Repeat Copy (blue).
Fader color denotes lower timestep in a sequence. Both can learn clusters
of hidden states yet NUTM exhibits clearer partition. \label{fig:Visualisation-of-the} }
\end{figure}

As previously mentioned in Sec. 3.3, MANN should let its states form
clusters to well-simulate Turing Machine. Fig. \ref{fig:Visualisation-of-the}
(a) and (c) show NTM actually organizes its $c_{t}$ space into clusters
corresponding to processing states (e.g, encoding and decoding). NUTM,
which explicitly partitions this space, clearly learn better clusters
of $c_{t}$ (see Fig. \ref{fig:Visualisation-of-the} (b) and (d)).
This contributes to NUTM's outperformance over NTM. 

\subsection{Program Usage Visualizations \label{subsec:Program-Usage-Visualizations}}

\ref{subsec:Visualization-on-program} and \ref{subsec:Visualization-on-program-1}
visualize the best inferences of NUTM on test data from single and
sequencing tasks. Each plot starts with the input sequence and the
predicted output sequence with error bits in the first row\footnote{Normally, black is bit 0, white is bit 1 in vector data. Orange is
prediction error. In tasks including priority sort, because data vectors
not only include value 0-1, but also other float values (e.g., priority
score), the color scale is automatically changed. Basically, error
bit is given darker color than 0 and lighter color than 1. For example,
in priority sort task, yellow is prediction error, and orange is bit
1.}. The second and fourth rows depict the read and write locations on
data memory, respectively. The third and fifth rows depict the program
distribution of the read head and write head, respectively. \ref{subsec:Perseveration-phenomenon-in}
visualizes random failed predictions of NTM on sequencing tasks. The
plots follow previous pattern except for the program distribution
rows. 

\subsubsection{Visualization on program distribution across timesteps (single tasks)\label{subsec:Visualization-on-program}}

\begin{figure}[H]
\begin{centering}
\includegraphics[width=1\linewidth]{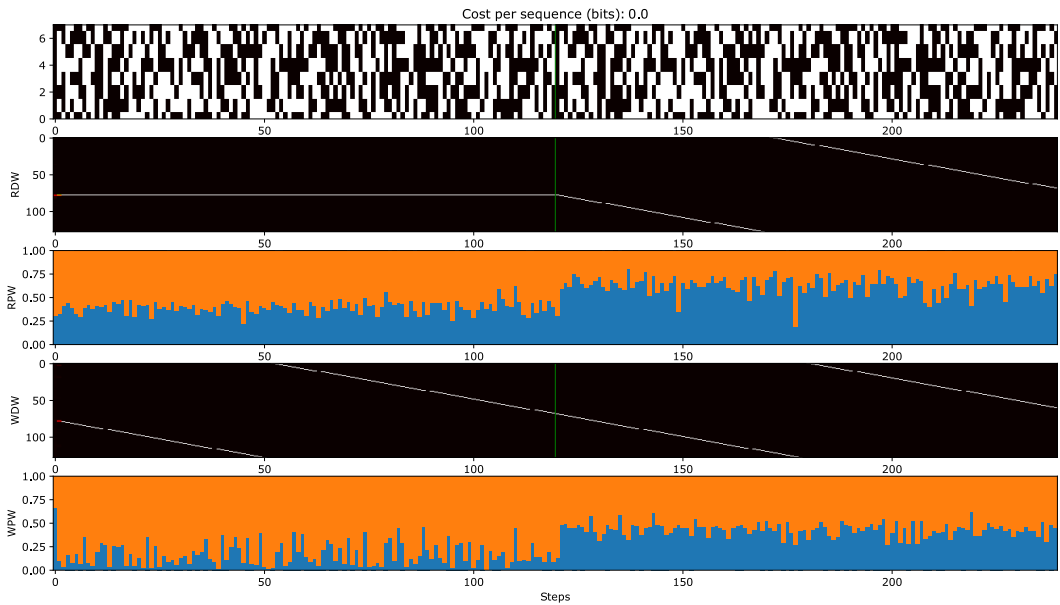}
\par\end{centering}
\caption{Copy (p=2).}

\end{figure}
\begin{figure}[H]
\begin{centering}
\includegraphics[width=1\linewidth]{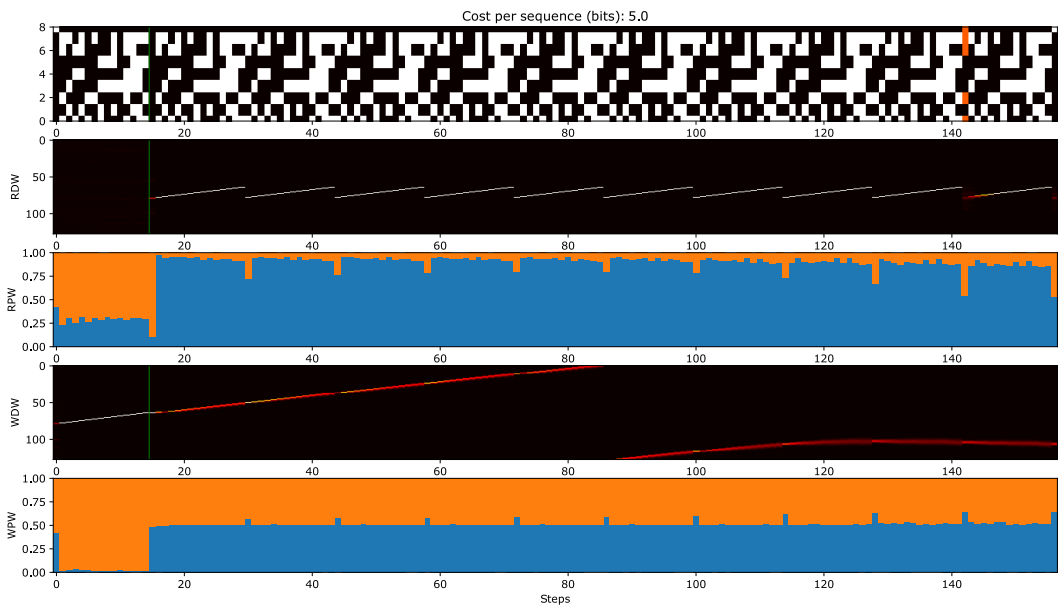}
\par\end{centering}
\caption{Repeat Copy (p=2).}
\end{figure}
\begin{figure}[H]
\begin{centering}
\includegraphics[width=1\linewidth]{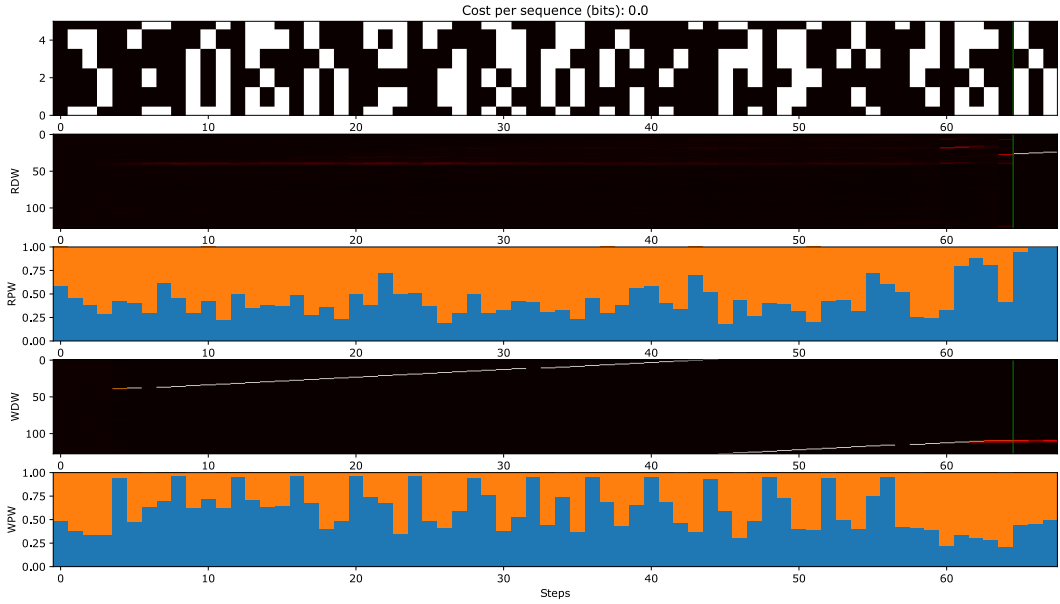}
\par\end{centering}
\caption{Associative Recall (p=2).}
\end{figure}
\begin{figure}[H]
\begin{centering}
\includegraphics[width=1\linewidth]{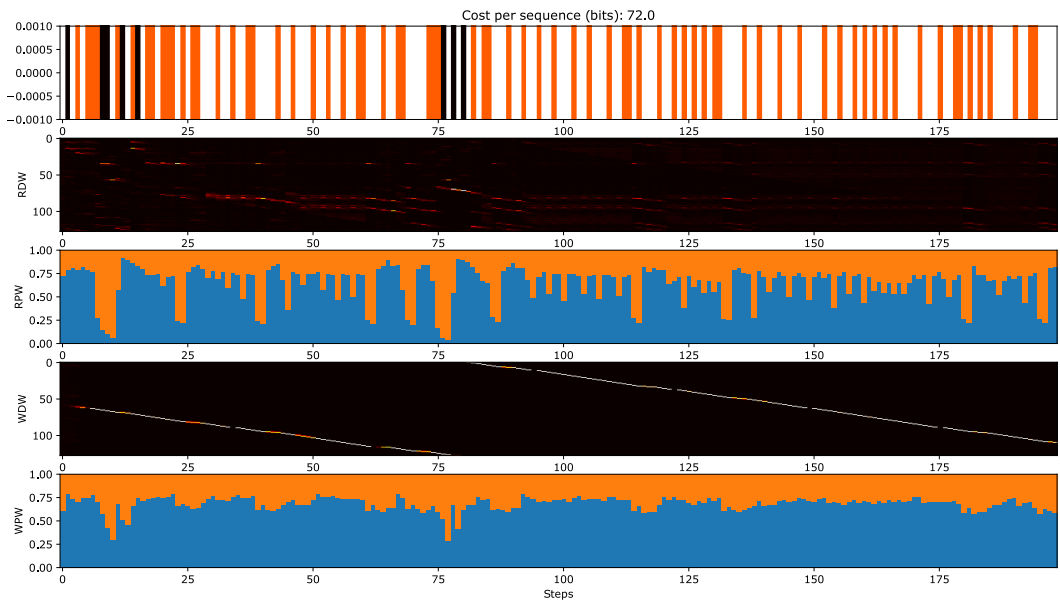}
\par\end{centering}
\caption{Dynamic N-grams (p=2).}
\end{figure}
\begin{figure}[H]
\begin{centering}
\includegraphics[width=1\linewidth]{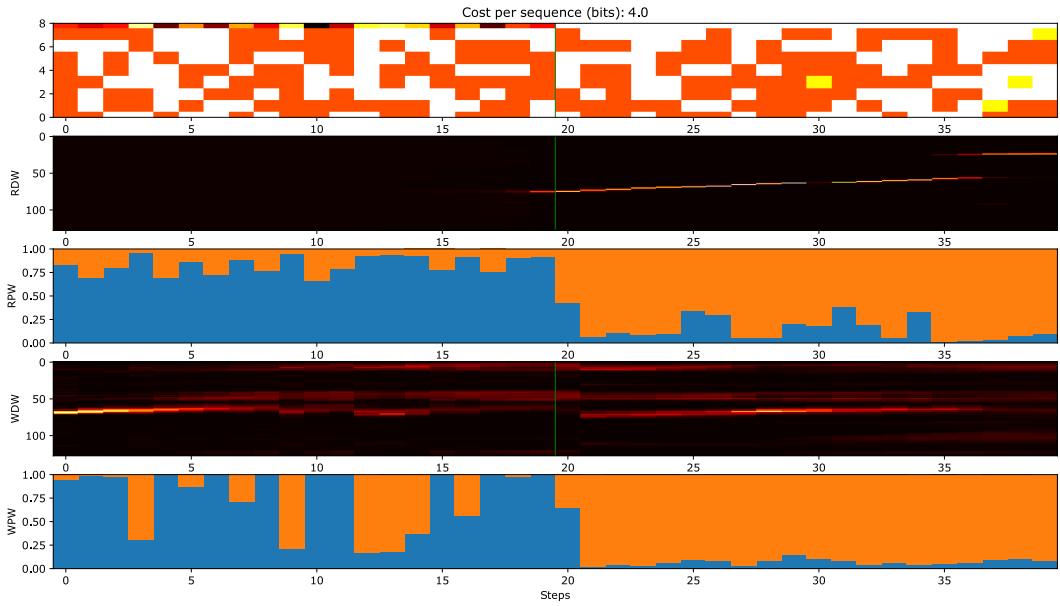}
\par\end{centering}
\caption{Priority Sort (p=2).}
\end{figure}
\begin{figure}[H]
\begin{centering}
\includegraphics[width=1\linewidth]{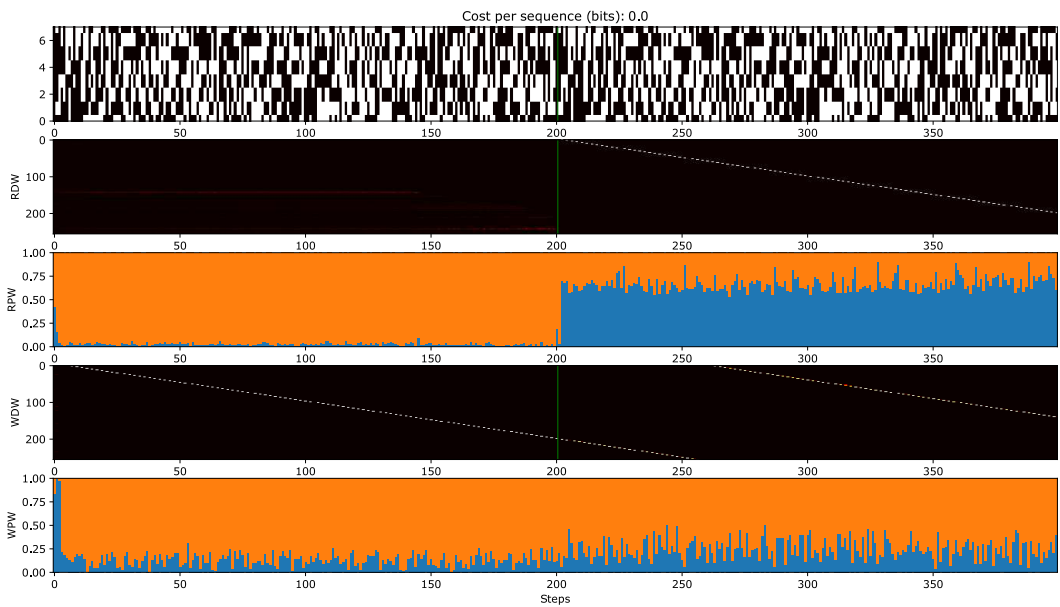}
\par\end{centering}
\caption{Long Copy (p=2).}
\end{figure}

\subsubsection{Visualization on program distribution across timesteps (sequencing
tasks)\label{subsec:Visualization-on-program-1}}

\begin{figure}[H]
\begin{centering}
\includegraphics[width=1\linewidth]{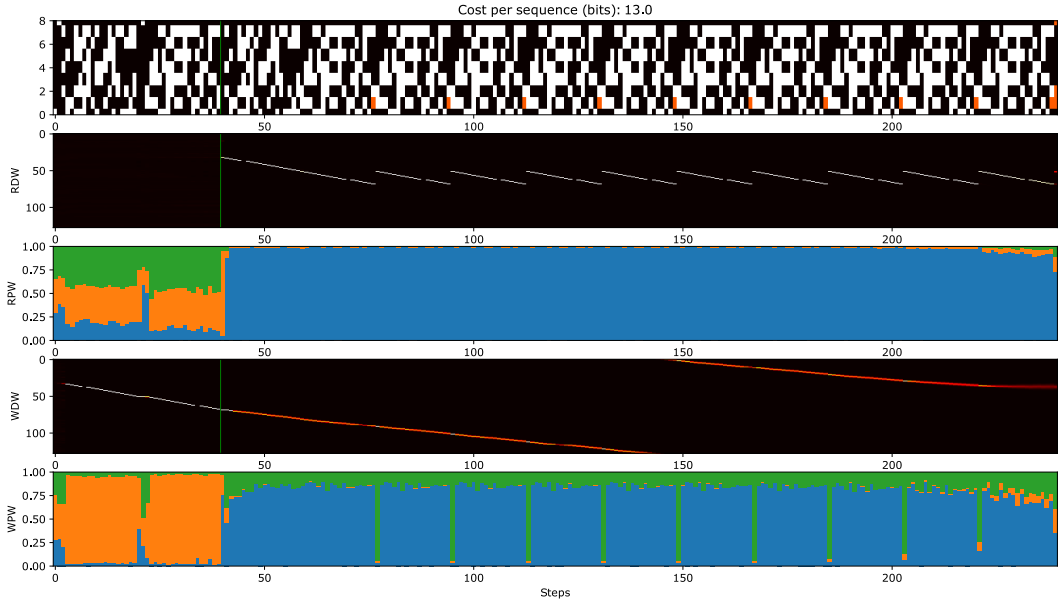}
\par\end{centering}
\caption{Copy+Repeat Copy (p=3).}
\end{figure}
\begin{figure}[H]
\begin{centering}
\includegraphics[width=1\linewidth]{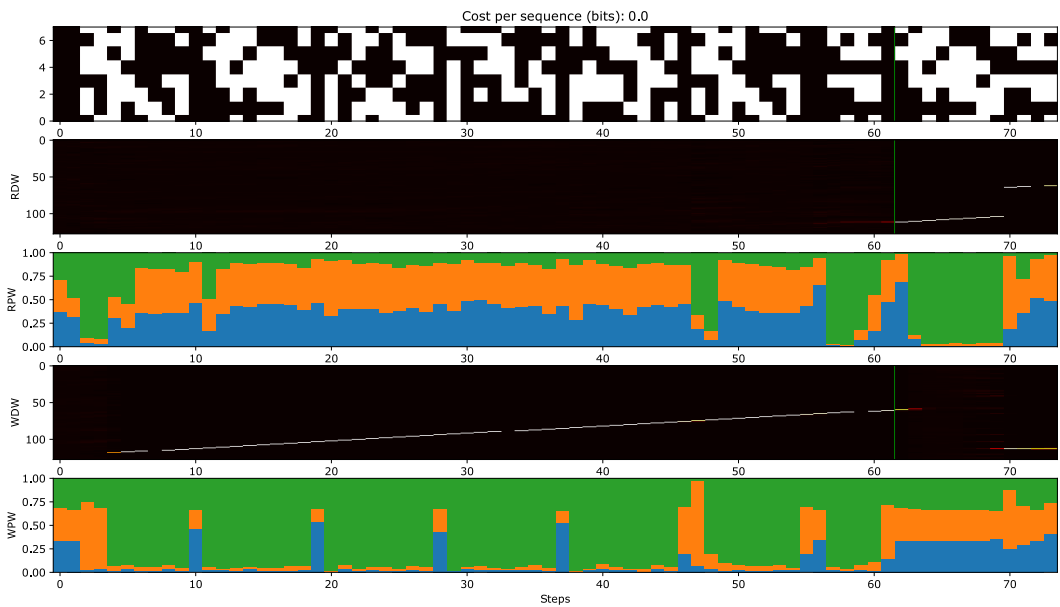}
\par\end{centering}
\caption{Copy+Associative Recall (p=3).}
\end{figure}
\begin{figure}[H]
\begin{centering}
\includegraphics[width=1\linewidth]{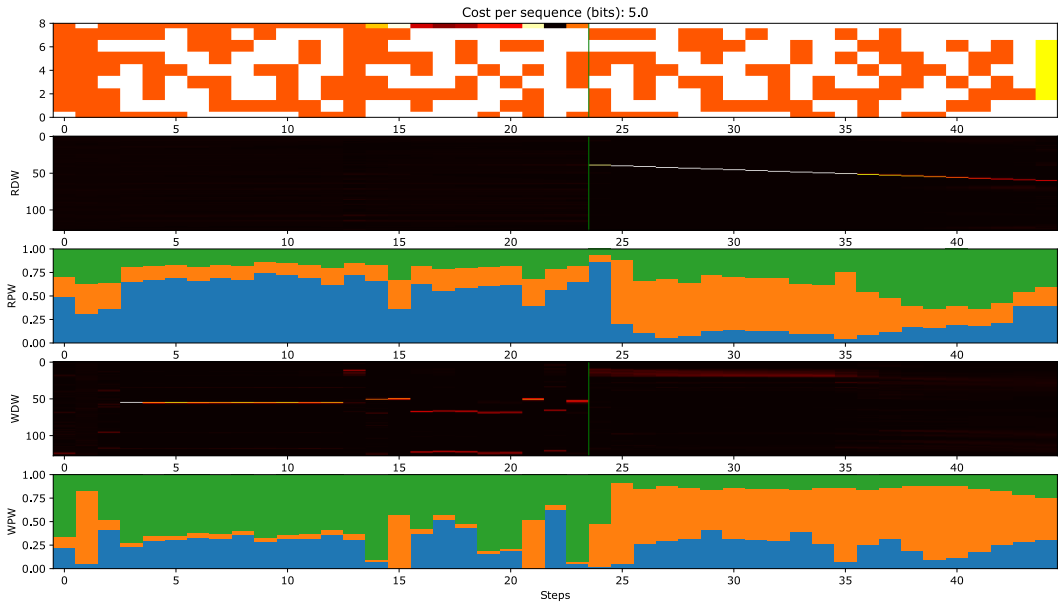}
\par\end{centering}
\caption{Copy+Priority Sort (p=3).}
\end{figure}
\begin{figure}[H]
\begin{centering}
\includegraphics[width=1\linewidth]{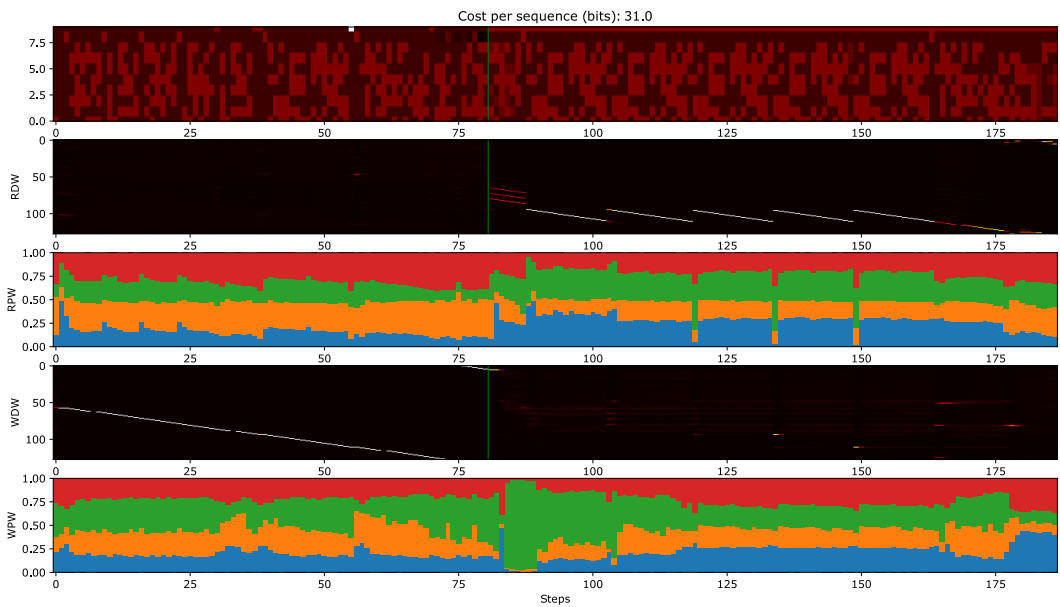}
\par\end{centering}
\caption{Copy+Repeat Copy+Associative Recall+Priority Sort (p=4).}
\end{figure}

\subsubsection{Perseveration phenomenon in NTM (sequencing tasks)\label{subsec:Perseveration-phenomenon-in}}

\begin{figure}[H]
\begin{centering}
\includegraphics[width=1\linewidth]{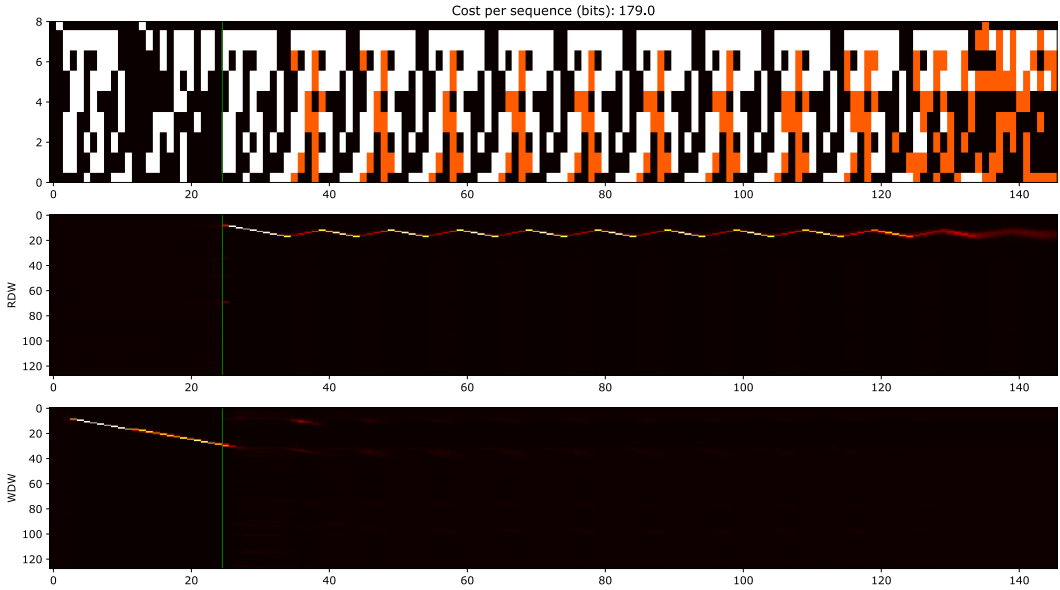}
\par\end{centering}
\caption{Copy+Repeat Copy perseveration (only Repeat Copy).}
\end{figure}
\begin{figure}[H]
\begin{centering}
\includegraphics[width=1\linewidth]{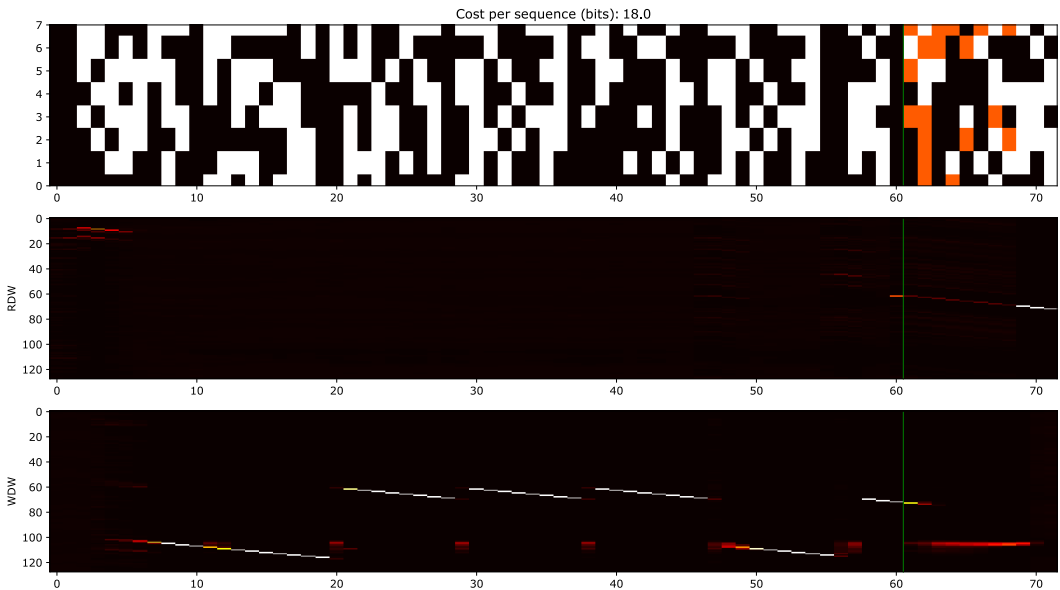}
\par\end{centering}
\caption{Copy+Associative Recall perseveration (only Copy).}
\end{figure}
\begin{figure}[H]
\begin{centering}
\includegraphics[width=1\linewidth]{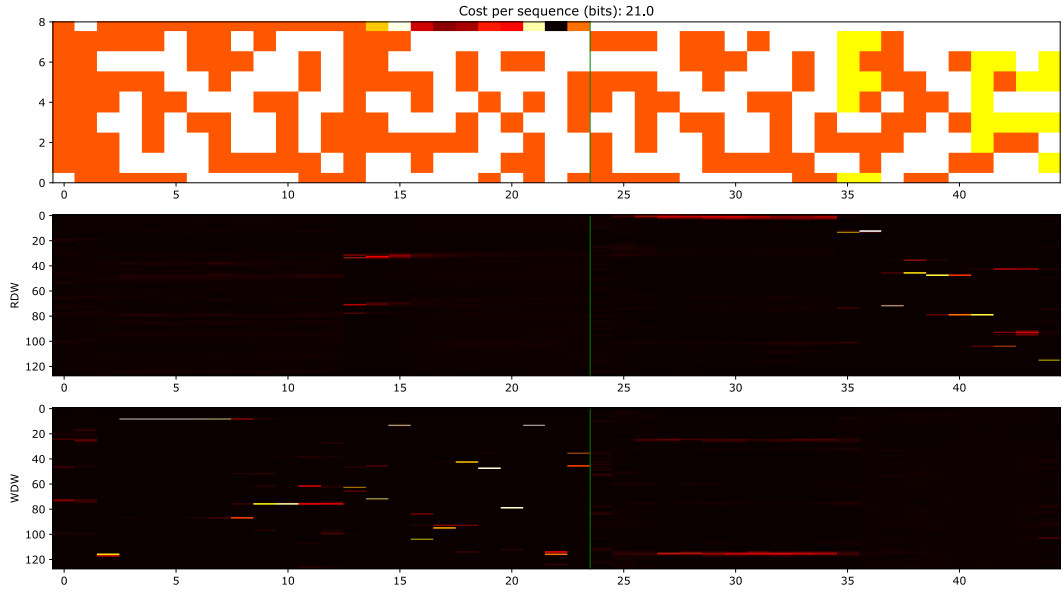}
\par\end{centering}
\caption{Copy+Priority Sort perseveration (only Copy).}
\end{figure}
\begin{figure}[H]
\begin{centering}
\includegraphics[width=1\linewidth]{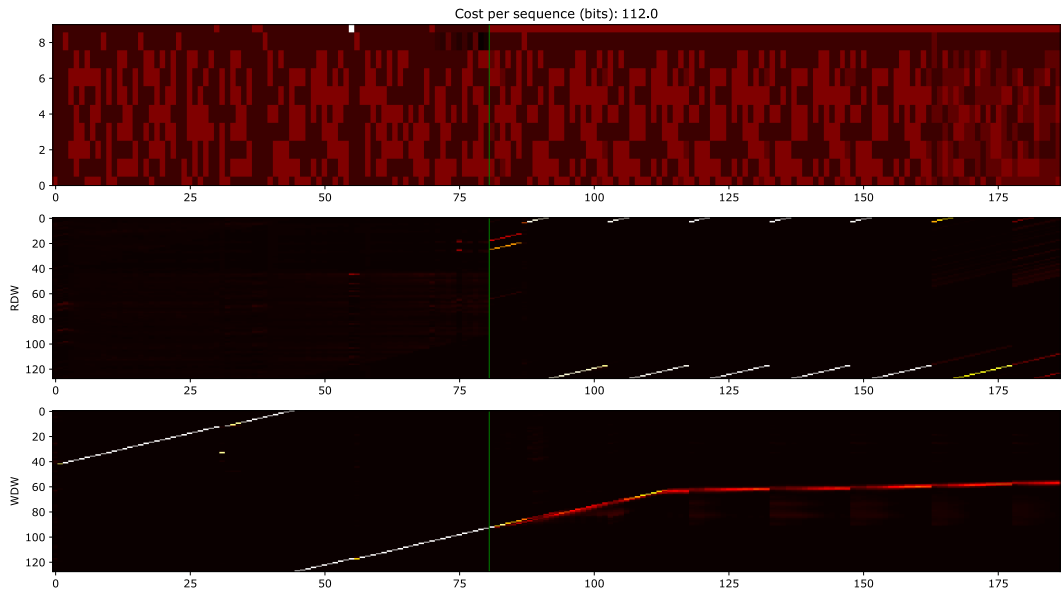}
\par\end{centering}
\caption{Copy+Repeat Copy+Associative Recall+Priority Sort perseveration (only
Repeat Copy).}
\end{figure}

\subsection{Details on Synthetic Tasks\label{subsec:Details-on-Synthetic}}

\subsubsection{NTM single tasks}

\begin{table}[H]
\begin{centering}
\begin{tabular}{ccccccccc}
\hline 
\multirow{2}{*}{Tasks} & \multicolumn{2}{c}{\#Read/Write Head\tablefootnote{In NTM, the number of read and write heads are equal.}} & \multicolumn{2}{c}{Controller Size} & \multicolumn{2}{c}{Memory Size} & \multicolumn{2}{c}{\#Parameters}\tabularnewline
\cline{2-9} 
 & NTM & NUTM & NTM & NUTM & NTM & NUTM & NTM & NUTM\tabularnewline
\hline 
Copy & 1 & 1 & 100 & 80 & 128 & 128 & 63,260 & 52,206\tabularnewline
\hline 
Repeat Copy & 1 & 1 & 100 & 80 & 128 & 128 & 63,381 & 52,307\tabularnewline
\hline 
Associative Recall & 1 & 1 & 100 & 80 & 128 & 128 & 62,218 & 51,364\tabularnewline
\hline 
Dynamic N-grams & 1 & 1 & 100 & 80 & 128 & 128 & 58,813 & 48,619\tabularnewline
\hline 
Priority Sort & 5 & 5 & 200 & 150 & 128 & 128 & 344,068 & 302,398\tabularnewline
\hline 
Long Copy & 1 & 1 & 100 & 80 & 256 & 256 & 63,260 & 52,206\tabularnewline
\hline 
\end{tabular}
\par\end{centering}
~

\caption{Model hyper-parameters (single tasks).}

\end{table}
\begin{table}[H]
\begin{centering}
\begin{tabular}{lll}
\hline 
\multirow{1}{*}{Tasks} & Training & Testing\tabularnewline
\hline 
Copy & Sequence length range: {[}1, 20{]} & Sequence length: 120\tabularnewline
\hline 
\multirow{2}{*}{Repeat Copy} & Sequence length range: {[}1, 10{]} & Sequence length range: {[}10, 20{]}\tabularnewline
 & \#Repeat range: {[}1, 10{]} & \#Repeat range: {[}10, 20{]}\tabularnewline
\hline 
\multirow{2}{*}{Associative Recall} & \#Item range: {[}2, 6{]} & \#Item range: {[}6, 20{]}\tabularnewline
 & Item length: 3 & Item length: 3\tabularnewline
\hline 
Dynamic N-grams & Sequence length: 50 & Sequence length: 200\tabularnewline
\hline 
\multirow{2}{*}{Priority Sort} & \#Item: 20 & \#Item: 20\tabularnewline
 & \#Sorted Item: 16 & \#Sorted Item: 20\tabularnewline
\hline 
Long Copy & Sequence length range: {[}1, 40{]} & Sequence length: 200\tabularnewline
\hline 
\end{tabular}
\par\end{centering}
~

\caption{Task settings (single tasks).}
\end{table}

\subsubsection{NTM sequencing tasks}

\begin{table}[H]
\begin{centering}
\begin{tabular}{ccccccccc}
\hline 
\multirow{2}{*}{Tasks} & \multicolumn{2}{c}{\#Read/Write Head} & \multicolumn{2}{c}{Controller Size} & \multicolumn{2}{c}{Memory Size} & \multicolumn{2}{c}{\#Parameters}\tabularnewline
\cline{2-9} 
 & NTM & NUTM & NTM & NUTM & NTM & NUTM & NTM & NUTM\tabularnewline
\hline 
C+RC & 1 & 1 & 200 & 150 & 128 & 128 & 206,481 & 153,941\tabularnewline
\hline 
C+AR & 1 & 1 & 200 & 150 & 128 & 128 & 206,260 & 153,770\tabularnewline
\hline 
C+PS & 3 & 3 & 200 & 150 & 128 & 128 & 275,564 & 263,894\tabularnewline
\hline 
C+RC+AR+PS & 3 & 3 & 250 & 200 & 128 & 128 & 394,575 & 448,379\tabularnewline
\hline 
\end{tabular}
\par\end{centering}
~

\caption{Model hyper-parameters (sequencing tasks).}
\end{table}
\begin{table}[H]
\begin{centering}
\begin{tabular}{lll}
\hline 
\multirow{1}{*}{Tasks} & Training & Testing\tabularnewline
\hline 
\multirow{2}{*}{C+RC} & Sequence length range: {[}1, 10{]} & Sequence length range: {[}10, 20{]}\tabularnewline
 & \#Repeat range: {[}1, 10{]} & \#Repeat range: {[}10, 15{]}\tabularnewline
\hline 
\multirow{3}{*}{C+AR} & Sequence length range: {[}1, 10{]} & Sequence length range: {[}10, 20{]}\tabularnewline
 & \#Item range: {[}2, 4{]} & \#Item range: {[}4, 6{]}\tabularnewline
 & Item length: 8 & Item length: 8\tabularnewline
\hline 
\multirow{3}{*}{C+PS} & Sequence length range: {[}1, 10{]} & Sequence length range: {[}10, 20{]}\tabularnewline
 & \#Item: 10 & \#Item: 10\tabularnewline
 & \#Sorted Item: 8 & \#Sorted Item: 10\tabularnewline
\hline 
\multirow{6}{*}{C+RC+AR+PS} & Sequence length range: {[}1, 10{]} & Sequence length range: {[}10, 20{]}\tabularnewline
 & \#Repeat range: {[}1, 5{]} & \#Repeat: 6\tabularnewline
 & \#Item range: {[}2, 4{]} & \#Item: 5\tabularnewline
 & Item length: 6 & Item length: 6\tabularnewline
 & \#Item: 10 & \#Item: 10\tabularnewline
 & \#Sorted Item: 8 & \#Sorted Item: 10\tabularnewline
\hline 
\end{tabular}
\par\end{centering}
~

\caption{Task settings (sequencing tasks).}
\end{table}

\subsubsection{Continual procedure learning tasks}

\begin{table}[H]
\begin{centering}
\begin{tabular}{cccccccc}
\hline 
\multicolumn{2}{c}{\#Read/Write Head} & \multicolumn{2}{c}{Controller Size} & \multicolumn{2}{c}{Memory Size} & \multicolumn{2}{c}{\#Parameters}\tabularnewline
\hline 
NTM & NUTM & NTM & NUTM & NTM & NUTM & NTM & NUTM\tabularnewline
\hline 
1 & 1 & 200 & 150 & 128 & 128 & 206,444 & 196,590\tabularnewline
\hline 
\end{tabular}
\par\end{centering}
~

\caption{Model hyper-parameters (continual procedure learning tasks). NUTM
uses 6 programs per head.}
\end{table}
\begin{table}[H]
\begin{centering}
\begin{tabular}{lll}
\hline 
\multirow{1}{*}{Tasks} & Training & Testing\tabularnewline
\hline 
Copy & Sequence length range: {[}1, 10{]} & Sequence length range: {[}1, 10{]}\tabularnewline
\hline 
\multirow{2}{*}{Repeat Copy} & Sequence length range: {[}1, 5{]} & Sequence length range: {[}1, 5{]}\tabularnewline
 & \#Repeat range: {[}1, 5{]} & \#Repeat range: {[}1, 5{]}\tabularnewline
\hline 
\multirow{3}{*}{Associative Recall} & Sequence length: 3 & Sequence length: 3\tabularnewline
 & \#Item range: {[}2, 3{]} & \#Item range: {[}2, 3{]}\tabularnewline
 & Item length: 3 & Item length: 3\tabularnewline
\hline 
\multirow{2}{*}{Priority Sort} & \#Item: 10 & \#Item: 10\tabularnewline
 & \#Sorted Item: 8 & \#Sorted Item: 8\tabularnewline
\hline 
\end{tabular}
\par\end{centering}
~

\caption{Task settings (continual procedure learning tasks).}
\end{table}

\subsection{Details on Few-shot Learning Task\label{subsec:Details-on-Few-shot}}

We use similar hyper-parameters as in \citet{santoro2016meta} , which
are reported in Tab. \ref{tab:NUTM-hyper-parameters-for-1}.

\begin{table}[H]
\begin{centering}
\begin{tabular}{cccccccc}
\hline 
Model & $p$ & \#Read Head & \#Write Head & Controller Size & $N$ & $M$ & $\mathbf{M}_{p}.K$ Size\tabularnewline
\hline 
MANN (LRUA) & 1 & 4 & 1 & 200 & 128 & 40 & 0\tabularnewline
\hline 
NUTM (LRUA) & 2 & 4 & 1 & 180 & 128 & 40 & 2\tabularnewline
\hline 
NUTM (LRUA) & 3 & 4 & 1 & 150 & 128 & 40 & 3\tabularnewline
\hline 
\end{tabular}
\par\end{centering}
~

\caption{Hyper-parameters for few-shot learning. All models use RMSprop optimizer
with learning rate $10^{-4}$.\label{tab:NUTM-hyper-parameters-for-1}}
\end{table}
Testing accuracy through time is listed below,

\begin{figure}[H]
\begin{centering}
\includegraphics[width=1\linewidth]{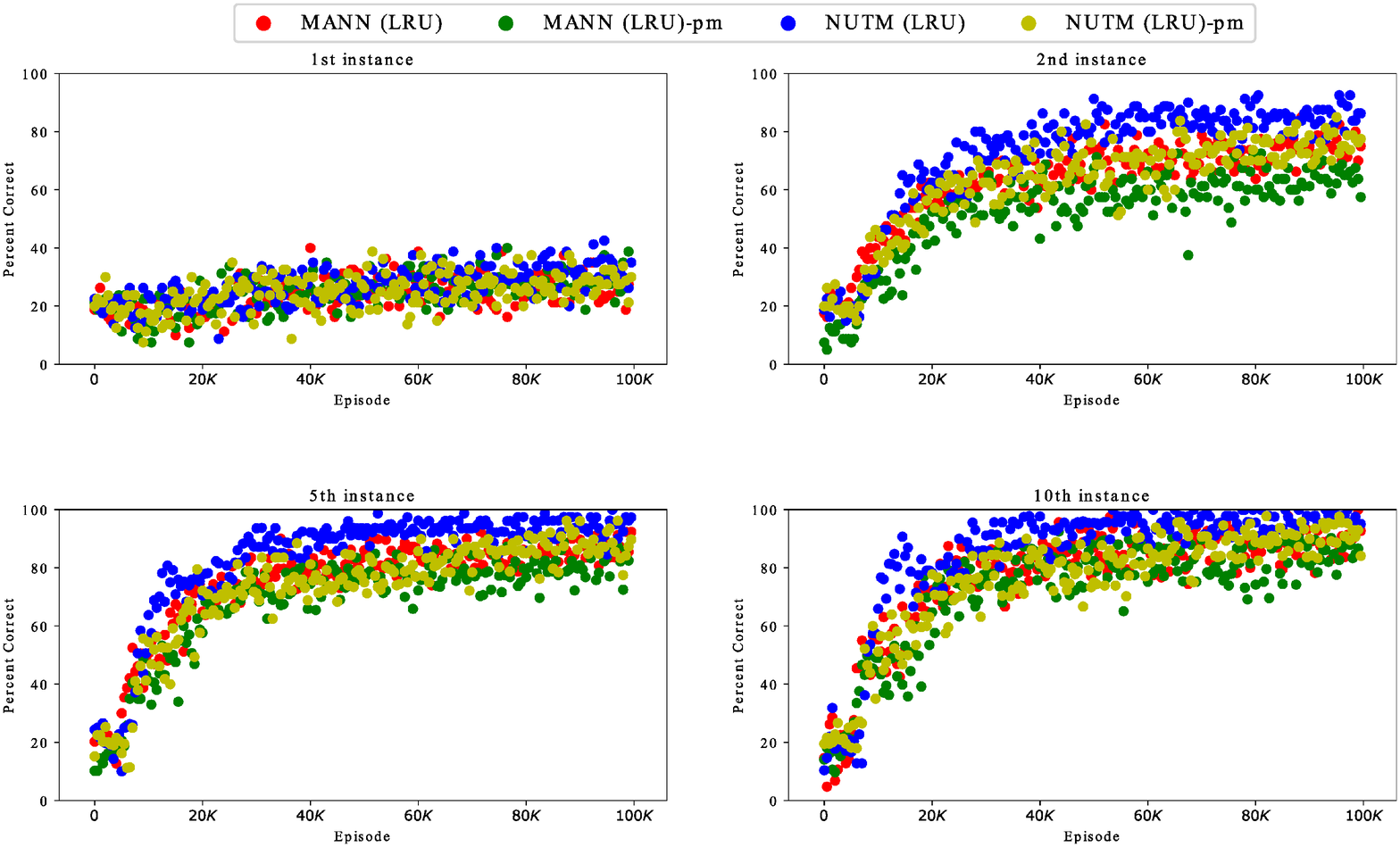}
\par\end{centering}
\caption{Testing accuracy during training (five random classes/episode, one-hot
vector labels, of length 50).}
\end{figure}
\begin{figure}[H]
\begin{centering}
\includegraphics[width=1\linewidth]{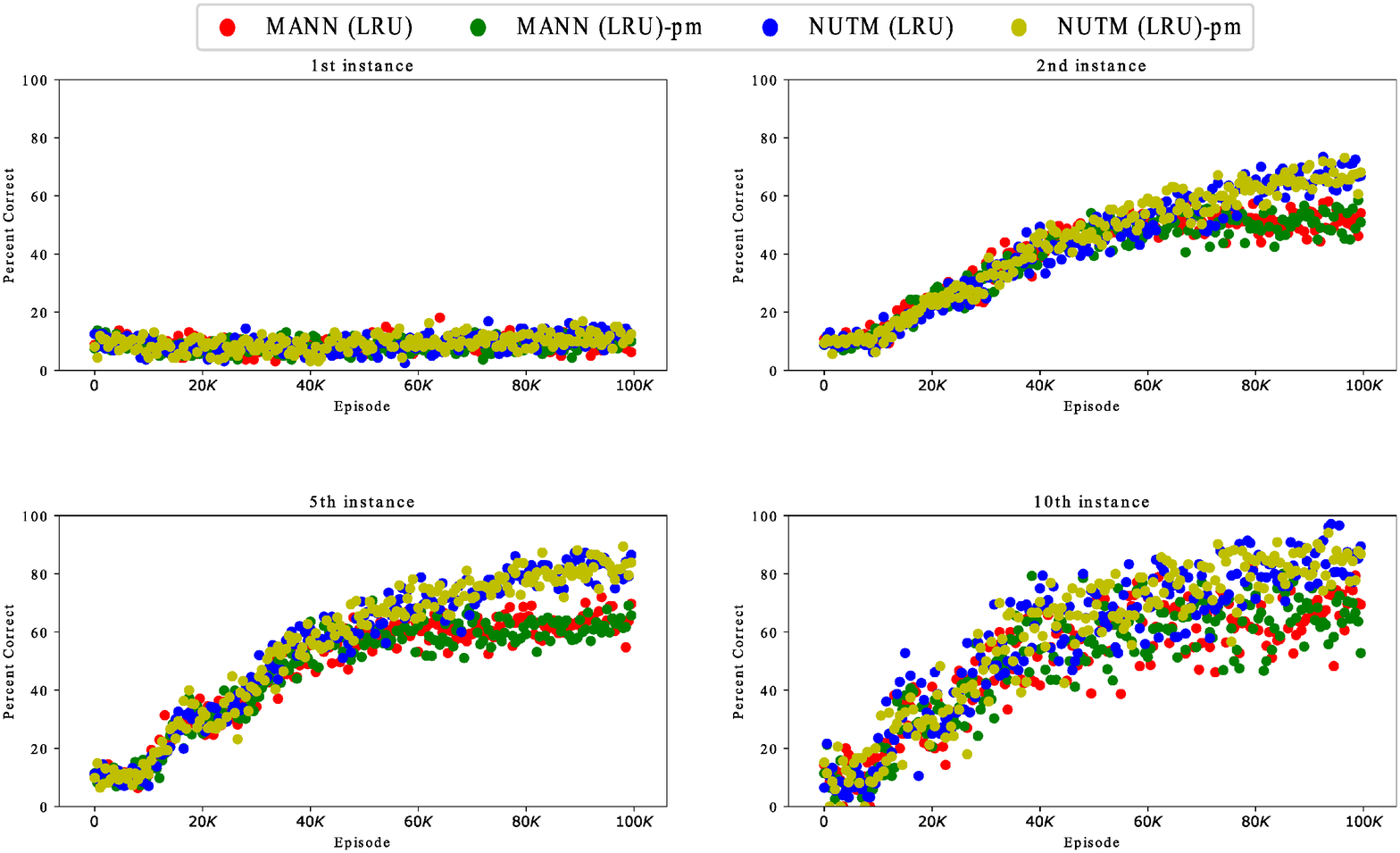}
\par\end{centering}
\caption{Testing accuracy during training (ten random classes/episode, one-hot
vector labels, of length 75).}
\end{figure}
\begin{table}
\begin{centering}
\begin{tabular}{l|c|ccc|ccc}
\hline 
\multirow{2}{*}{Model} & Persistent & \multicolumn{3}{c|}{5 classes} & \multicolumn{3}{c}{10 classes}\tabularnewline
\cline{3-8} 
 & memory\tablefootnote{If the memory is not artificially erased between episodes, it is called
persistent. This mode is hard for the case of 5 classes as shown in
\citet{santoro2016meta}} & $2^{nd}$ & $3^{rd}$ & $5^{th}$ & $2^{nd}$ & $3^{rd}$ & $5^{th}$\tabularnewline
\hline 
MANN (LRUA){*} & No & 82.8 & 91.0 & 94.9 & - & - & -\tabularnewline
MANN (LRUA) & No & 82.3 & 88.7 & 92.3 & 52.7 & 60.6 & 64.7\tabularnewline
NUTM (LRUA) & No & \textbf{85.7} & \textbf{91.3} & \textbf{95.5} & \textbf{68.0} & \textbf{78.1} & \textbf{82.8}\tabularnewline
\hline 
Human{*} & Yes & 57.3 & 70.1 & 81.4 & - & - & -\tabularnewline
MANN (LRUA){*} & Yes & $\approx58.0$ & - & $\approx75.0$ & $\approx60.0$ & - & $\approx80.0$\tabularnewline
MANN (LRUA) & Yes & 66.2 & 73.4 & 81.0 & 51.3 & 59.2 & 63.3\tabularnewline
NUTM (LRUA) & Yes & \textbf{77.8} & \textbf{85.8} & \textbf{89.8} & \textbf{69.0} & \textbf{77.9} & \textbf{82.7}\tabularnewline
\hline 
\end{tabular}
\par\end{centering}
~

\caption{Test-set classification accuracy (\%) on the Omniglot dataset after
100,000 episodes of training. {*} denotes available results from \citet{santoro2016meta}
(some are estimated from plotted figures).\label{tab:meta}}
\end{table}

\subsection{Details on bAbI Task\label{subsec:Details-on-bAbI}}

We train the models using RMSprop optimizer with fixed learning rate
of $10^{-4}$ and momentum of 0.9. The batch size is 32 and we adopt
layer normalization \cite{lei2016layer} to DNC's layers. Following
\citet{W18-2606} practice, we also remove temporal linkage for faster
training. The details of hyper-parameters are listed in Table \ref{tab:NUTM-hyper-parameters-for}.
Full NUTM ($p=4$) results are reported in Table \ref{tab:babifull}.

\begin{table}[H]
\begin{centering}
\begin{tabular}{cccccccc}
\hline 
\#Read Head & \#Write Head & Controller Size & $N$ & $M$ & $p$ & $\mathbf{M}_{p}.K$ Size & \#Parameters\tabularnewline
\hline 
4 & 1 & 256 & 196 & 64 & 1\tablefootnote{When $p=1$, the model converges to layer-normed DNC} & 1 & 891,136\tabularnewline
\hline 
4 & 1 & 200 & 196 & 64 & 2 & 2 & 934,787\tabularnewline
\hline 
4 & 1 & 172 & 196 & 64 & 4 & 4 & 794,773\tabularnewline
\hline 
\end{tabular}
\par\end{centering}
~

\caption{NUTM hyper-parameters for bAbI.\label{tab:NUTM-hyper-parameters-for}}
\end{table}
\begin{table}[H]
\begin{centering}
\begin{tabular}{lcc}
\hline 
Task & bAbI Best Results & bAbI Mean Results\tabularnewline
\hline 
1: 1 supporting fact & 0.0 & 0.0 $\pm$ 0.0\tabularnewline
2: 2 supporting facts  & 0.2 & 0.6 $\pm$ 0.3\tabularnewline
3: 3 supporting facts  & 4.0 & 7.6 $\pm$ 3.9\tabularnewline
4: 2 argument relations  & 0.0 & 0.0 $\pm$ 0.0\tabularnewline
5: 3 argument relations  & 0.4 & 1.0 $\pm$ 0.4\tabularnewline
6: yes/no questions  & 0.0 & 0.0 $\pm$ 0.0\tabularnewline
7: counting  & 1.9 & 1.5 $\pm$ 0.8\tabularnewline
8: lists/sets  & 0.6 & 0.3 $\pm$ 0.2\tabularnewline
9: simple negation  & 0.0 & 0.0 $\pm$ 0.0\tabularnewline
10: indefinite knowledge  & 0.1 & 0.1 $\pm$ 0.0\tabularnewline
11: basic coreference  & 0.0 & 0.0 $\pm$ 0.0\tabularnewline
12: conjunction  & 0.0 & 0.0 $\pm$ 0.0\tabularnewline
13: compound coreference  & 0.1 & 0.0 $\pm$ 0.0\tabularnewline
14: time reasoning  & 0.3 & 1.6 $\pm$ 2.2\tabularnewline
15: basic deduction  & 0.0 & 2.6 $\pm$ 8.3\tabularnewline
16: basic induction  & 49.3 & 52.0 $\pm$ 1.7\tabularnewline
17: positional reasoning  & 4.7 & 18.4 $\pm$ 12.7\tabularnewline
18: size reasoning  & 0.4 & 1.6 $\pm$ 1.1\tabularnewline
19: path finding  & 4.3 & 23.7 $\pm$ 32.2\tabularnewline
20: agent\textquoteright s motivation  & 0.0 & 0.0 $\pm$ 0.0\tabularnewline
\hline 
Mean Error (\%) & 3.3 & 5.6 $\pm$ 1.9\tabularnewline
\hline 
Failed (Err. \textgreater 5\%) & 1 & 3 $\pm$ 1.2\tabularnewline
\hline 
\end{tabular}
\par\end{centering}
~

\caption{NUTM ($p=4$) bAbI best and mean errors (\%).\label{tab:babifull}}
\end{table}

\subsection{Example of memory operation function in NTM\label{subsec:Example-of-memory}}

In NTM, $\xi_{t}=\left\{ \beta_{t},k_{t},g_{t},s_{t},e_{t},v_{t}\right\} $.
The memory addressing weight is initially computed by content-based
attention,

\begin{equation}
w_{t}^{c}\left(i\right)=\frac{\exp\left(\beta_{t}m\left(k_{t},M_{t}\left(i\right)\right)\right)}{\stackrel[j=1]{D}{\sum}\exp\left(\beta_{t}m\left(k_{t},M_{t}\left(j\right)\right)\right)}\label{eq:cbaseatt}
\end{equation}
Here, $w_{t}^{c}\in\ensuremath{\mathbb{R}}^{N}$ is the content-based
weight, $\beta_{t}$ is a strength scalar, and $m$ is implemented
as cosine similarity

\begin{equation}
m\left(k_{t},M_{t}(i)\right)=\frac{k_{t}\cdot M_{t}(i)}{||k_{t}||\cdot||M_{t}(i)||}
\end{equation}
In addition, NTM supports location-based addressing started with an
interpolation between content-based weight and the previous weight

\begin{equation}
w_{t}^{g}=g_{t}w_{t}^{c}+\left(1-g_{t}\right)w_{t}
\end{equation}
where $g_{t}$ is the interpolation gate that determines to use (or
ignore) content-based addressing. Then, NTM can shift the address
to other rows by performing convolution shift modulo $R$,

\begin{equation}
\tilde{w_{t}}\left(i\right)=\stackrel[j=0]{R}{\sum}w_{t}^{g}\left(i\right)s_{t}\left(i-j\right)
\end{equation}
where $s_{t}$ is the shift weighting. To prevent the shifted weight
from blurring, sharpening is applied

\begin{equation}
w_{t}\left(i\right)=\frac{\tilde{w_{t}}\left(i\right)^{\gamma}}{\underset{j}{\sum}\tilde{w_{t}}\left(j\right)^{\gamma}}
\end{equation}
Then, the memory is updated as follows,

\begin{align}
M_{t}^{erased}\left(i\right) & =M_{t-1}\left(i\right)\left[1-w_{t}\left(i\right)e_{t}\right]\\
M_{t}\left(i\right) & =M_{t}^{erased}\left(i\right)+w_{t}\left(i\right)v_{t}
\end{align}
where $e_{t}\in\ensuremath{\mathbb{R}}^{D}$ and $v_{t}\in\ensuremath{\mathbb{R}}^{D}$
are erase vector and update vector, respectively. The read value is
computed using the same address weight as follows,

\begin{equation}
r=\stackrel[i=1]{N}{\sum}w_{t}\left(i\right)M_{t}\left(i\right)
\end{equation}

\subsection{Others}

If we deliberately set the key dimension equal to the number of programs,
we can even place an orthogonal basis constraint on the key space
of NSM by minimizing the following loss, 

\begin{equation}
l_{p_{2}}=\left\Vert \mathbf{M}_{p}.K\mathbf{M}_{p}.K^{T}-\mathbf{I}\right\Vert 
\end{equation}
where $\mathbf{M}_{p}.K$ and $\mathbf{I}$ denote the key part in
NSM and the identity matrix, respectively. 

Direct attention is one special case of key-value attention when the
memory keys form orthogonal basis. When this happens, the generated
key $k_{t}^{p}$ plays a direct role as the attention weight $w_{t}^{p}$.
Thus, using key-value attention is more generic.

For all tasks, $\eta_{t}$ is fixed to $0.1$, reducing with decay
rate of $0.9$. 

\begin{figure}

\begin{centering}
\includegraphics[width=1\textwidth]{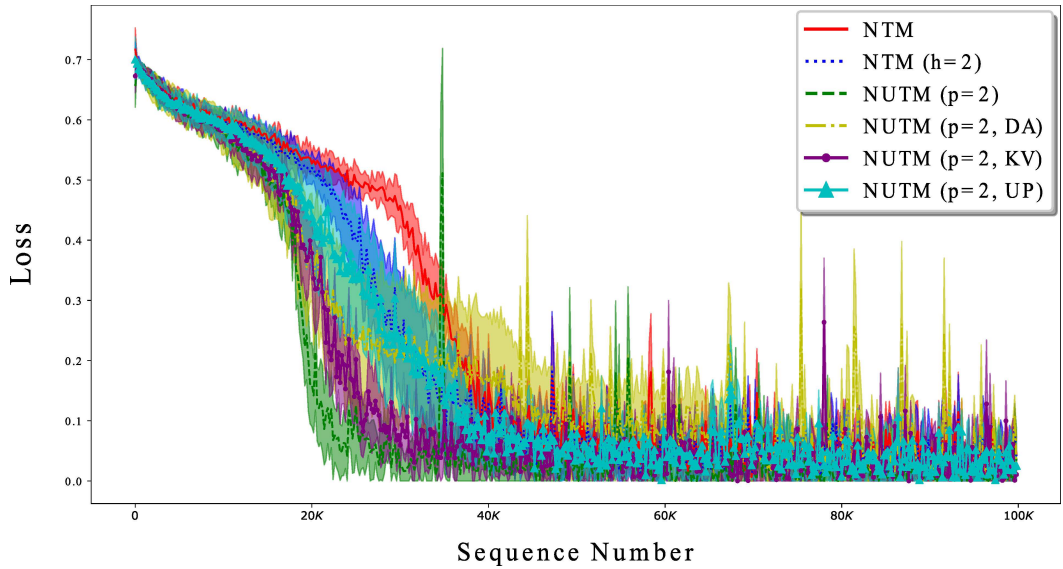}
\par\end{centering}
\caption{Learning curves on Associative Recall (AR) ablation study.\label{fig:Learning-curves-on-2}}

\end{figure}

Ablation study's learning losses with mean and error bar are plotted
in Fig. \ref{fig:Learning-curves-on-2}.

\end{document}